\documentclass[10pt]{article}

\usepackage[preprint]{tmlr}

\input{t1lmr.fd}
\DeclareFontShape{T1}{lmr}{bx}{sc}{<->ssub*lmr/m/sc}{}
\input{t1lmss.fd}
\DeclareFontShape{T1}{lmss}{bx}{sc}{<->ssub*lmss/m/n}{}

\usepackage{hyperref}
\hypersetup{hypertexnames=false}
\usepackage{url}
\usepackage{microtype}
\usepackage{inconsolata}

\def\month{MM}
\def\year{2026}
\def\openreview{\url{https://openreview.net/forum?id=XXXX}}

\usepackage{amsmath,amssymb,amsthm}
\usepackage{mathtools}
\usepackage{dsfont}  

\usepackage{booktabs}
\usepackage{tabularx}
\usepackage{graphicx}
\usepackage{wrapfig}
\usepackage{float}

\usepackage{listings}

\usepackage{algorithm}
\usepackage{algpseudocode}

\usepackage{enumitem}
\usepackage{tikz}
\usepackage{pgfplots}
\pgfplotsset{compat=1.18}
\usetikzlibrary{shapes,arrows,arrows.meta,positioning,fit,backgrounds,shadows.blur,decorations.pathreplacing,calc}

\definecolor{codegreen}{rgb}{0,0.6,0}
\definecolor{codegray}{rgb}{0.5,0.5,0.5}
\definecolor{codepurple}{rgb}{0.58,0,0.82}
\definecolor{backcolour}{rgb}{0.95,0.95,0.92}
\definecolor{hlalgo}{rgb}{0.92,0.95,1.0}

\lstdefinestyle{pythonstyle}{
    backgroundcolor=\color{backcolour},
    commentstyle=\color{codegreen},
    keywordstyle=\color{magenta},
    numberstyle=\tiny\color{codegray},
    stringstyle=\color{codepurple},
    basicstyle=\ttfamily\footnotesize,
    breakatwhitespace=false,
    breaklines=true,
    captionpos=b,
    keepspaces=true,
    numbers=left,
    numbersep=5pt,
    showspaces=false,
    showstringspaces=false,
    showtabs=false,
    tabsize=2
}
\lstdefinestyle{cudastyle}{
    backgroundcolor=\color{backcolour},
    basicstyle=\ttfamily\tiny,
    breaklines=true,
    keepspaces=true,
    numbers=none,
    showstringspaces=false,
    tabsize=4,
    language=C++,
    morekeywords={__global__,__device__,__shared__,__restrict__,__launch_bounds__,
                  torch,Tensor,float4,__nv_bfloat162,int64_t,dim3,cudaStream_t},
    commentstyle=\color{codegreen},
    keywordstyle=\color{magenta},
    stringstyle=\color{codepurple},
}

\definecolor{ifcoheader}{RGB}{44,62,80}

\theoremstyle{definition}

\title{Optimizing CUDA like a Human: Micro-Profiling Tools as Expert Surrogates for LLM-Based GPU Kernel Optimization}

\author{\name Jiading Gai\textsuperscript{*} \email jiadingg@amazon.com \\
      \addr Amazon Web Services
      \AND
      \name Shuai Zhang\textsuperscript{*} \email shuaizs@amazon.com \\
      \addr Amazon Web Services
      \AND
      \name Kaj Bostrom \email bostromk@amazon.com \\
      \addr Amazon Web Services
      \AND
      \name Jin Huang \email jinhun@amazon.com \\
      \addr Amazon Web Services
      \AND
      \name Vihang Patil \email pvihang@amazon.com \\
      \addr Amazon Web Services
      \AND
      \name Haoyang Fang \email haoyfang@amazon.com \\
      \addr Amazon Web Services
      \AND
      \name Bernie Wang \email yuyawang@amazon.com \\
      \addr Amazon Web Services
      \AND
      \name Huzefa Rangwala\textsuperscript{\dag} \email rhuzefa@gmail.com \\
      \addr Siemens
      \AND
      \name George Karypis\textsuperscript{\dag} \email karypis@umn.edu \\
      \addr University of Minnesota
}

\usepackage{xspace}
\usepackage[most]{tcolorbox}

\newcommand{\model}{\textsc{KernelPro}\xspace}

\begin{document}

\emergencystretch=3em
\hbadness=2500
\raggedbottom

\maketitle
\renewcommand{\thefootnote}{\fnsymbol{footnote}}
\footnotetext[1]{Equal contribution.\nobreak\hspace{1.5em}%
\textsuperscript{\dag}Work done while at AWS.}
\renewcommand{\thefootnote}{\arabic{footnote}}

\begin{abstract}
We present \model, a closed-loop multi-agent system that automatically generates, profiles, and iteratively optimizes GPU kernel code by integrating large language model (LLM) code generation with hardware profiler feedback and pluggable bottleneck detection tools. \model introduces four contributions: (1) a \emph{semantic feedback operator} that encodes expert heuristics as pluggable micro-profiling tools, transforming raw hardware metrics into actionable natural language guidance, (2) a \emph{two-stage tool invocation architecture} where roofline-based bottleneck classification filters which specialized analysis tools execute, combining kernel-level (\texttt{ncu}), instruction-level (SASS), and system-level (\texttt{nsys}) profiling, (3) a \emph{domain-adapted MCTS} with progressive widening, asymmetric branching, log-reward calibration, dead-end pruning, and search memory for cross-iteration learning, and (4) \emph{direct CuTe source-level code generation} via autonomous code search over the CUTLASS/CuTe codebase, mimicking how expert engineers write high-performance GPU kernels in raw CUDA+CuTe. On KernelBench, \model achieves geometric mean speedups of 2.42$\times$/4.69$\times$/5.30$\times$ on Levels 1/2/3, establishing state-of-the-art (SOTA) performance across all difficulty levels. On VeOmni's expert-optimized MoE training kernels, \model achieves 1.23$\times$ over hand-tuned Triton by generating a from-scratch raw-CUDA+CuTe Hopper WGMMA kernel. Ablation studies demonstrate that each design component independently and significantly improves optimization quality: micro-profiling tools ($p < 0.0001$ vs raw metrics), MCTS search (26\% higher geometric mean vs greedy, $p = 0.004$), and proactive tool orchestration (23\% improvement, $p = 0.035$). Finally, \model is the first CUDA kernel coding agent to optimize \emph{energy efficiency} beyond the speed-only focus of prior systems, demonstrating an 11.6\% measured energy reduction at matched speed.
\end{abstract}

\begin{center}
\textcolor{blue}{Code will be released upon publication.}
\end{center}


\section{Introduction}
\label{sec:intro}

Optimizing GPU kernels demands expertise that spans hardware architecture, memory hierarchies, instruction scheduling, and vendor-specific programming models. Experienced engineers follow a disciplined cycle: profile the kernel, recognize bottleneck patterns in the metrics, diagnose root causes, apply targeted transformations, and re-profile to verify. The bottleneck shifts after each successful optimization, and the process repeats until hardware limits are reached. What makes experts effective is not the ability to read numbers from a profiler, but the \emph{pattern recognition} and \emph{diagnostic reasoning} they apply to those numbers--heuristics refined over years and often codified into personal scripts and tools~\citep{nvidia2024bestpractices,yang2020hierarchical}.

Recent LLM-based CUDA optimization systems~\citep{zhang2025cudaforge,dong2026kernelblaster,li2026stitchcuda,chen2025cupilot} have shown that language models can generate competitive kernel code when given profiling data. However, these systems present raw or lightly summarized metrics directly to the LLM and rely on the model to implicitly replicate expert reasoning. This conflates two distinct capabilities: \emph{interpreting hardware telemetry} (a structured, rule-governed task) and \emph{generating optimized code} (a creative, context-dependent task). By asking the LLM to do both at once, existing agents forgo the systematic intermediate analysis that makes human experts reliable--they miss optimizations, produce inconsistent diagnoses, and cannot easily incorporate new analysis patterns without retraining.

We introduce \model, a closed-loop multi-agent system that separates these concerns by encoding expert heuristics as \emph{pluggable micro-profiling tools}--executable analysis functions that transform raw hardware metrics into actionable natural language guidance before the code-generation LLM sees them. Each tool implements a specific diagnostic pattern: a trigger condition (e.g., tensor core utilization below 10\%), analysis logic (e.g., check whether the kernel performs reducible matrix operations), and a prescriptive recommendation (e.g., ``rewrite using CUTLASS GEMM for a major expected speedup''). A two-stage invocation pipeline first classifies the kernel's bottleneck type via roofline analysis, then dispatches only the relevant tools--reducing prompt noise and focusing the LLM on pertinent optimizations.

\model further contributes a domain-adapted Monte Carlo Tree Search (MCTS) strategy that balances exploitation of promising optimization paths with exploration of alternative approaches. Unlike the flat iterative refinement used by prior LLM-based CUDA optimization systems, MCTS with UCT selection avoids premature convergence to local optima on complex multi-step optimization problems. Combined with a search memory mechanism that distills findings across iterations, \model achieves monotonically improving performance curves under fixed compute budgets.

On the KernelBench benchmark~\citep{ouyang2024kernelbench}, \model achieves geometric mean speedups of $2.42\times$/$4.69\times$/$5.30\times$ on Levels 1/2/3, surpassing KernelBlaster~\citep{dong2026kernelblaster}, the prior SOTA at time of writing, on all levels. Following the principle that design choices should be empirically established rather than assumed, we validate each of \model's components through controlled ablations---each assessed for significance with a paired Wilcoxon signed-rank test~\citep{hollander2014nonparametric}---on a fixed 42-task KernelBench subset (Levels 1--3), selected for transformer relevance and used throughout this paper (Section~\ref{sec:component-ablation}). For instance, micro-profiling tools provide 125\% higher speedup than raw metrics ($p < 0.0001$), and MCTS significantly outperforms greedy search (Wilcoxon $p = 0.004$, 26\% higher geometric mean).

Overall, \model contributes and synthesizes five ideas to make LLM-based optimization more grounded, interpretable, and effective on modern GPU architectures:
\begin{enumerate}[leftmargin=*,itemsep=2pt,topsep=2pt]
    \item A \textbf{semantic feedback operator} that formalizes expert heuristics as pluggable, LLM-invocable micro-profiling tools, transforming hardware metrics into natural language optimization guidance.

    \item A \textbf{two-stage tool invocation architecture} where roofline-based bottleneck classification (Stage~1) filters which specialized analysis tools execute (Stage~2), combining kernel-level, instruction-level, and system-level profiling.

    \item A \textbf{domain-adapted MCTS} for LLM-based CUDA optimization with progressive widening, asymmetric branching, log-reward calibration, and dead-end pruning. Search memory enables cross-iteration learning.

    \item \textbf{Direct CuTe source-level code generation} via autonomous code search over the CUTLASS/CuTe codebase~\citep{nvidia_cutlass} (CUTLASS is a CUDA template library of high-performance kernels; CuTe is its underlying tensor/layout-algebra layer for composing kernels from scratch), mimicking how expert engineers write high-performance GPU kernels in raw CUDA+CuTe (full generated source in Appendix~\ref{appendix:veomni-dw1-source}).

    \item The \textbf{first energy-aware CUDA kernel coding agent}: whereas prior LLM-based kernel agents optimize for speed alone, \model extends its reward and profiling tools to \emph{energy efficiency} as a secondary objective via a lexicographic energy-aware reward. As a preliminary study, a matched-speed A/B finds a measured $11.6\%$ energy reduction at identical speed via lower-energy instruction selection (Section~\ref{sec:energy-extension}); a full evaluation is left to future work.
\end{enumerate}

\section{Related Work}
\label{sec:related-work}

Several recent systems apply LLMs to CUDA kernel optimization, differing primarily in how hardware profiling data reaches the LLM. Some rely solely on end-to-end timing without hardware profiling~\citep{lange2025gpukernelscientist,zhang2026cudaagent,wiedemann2026kernelfoundry,chen2026avo}. Others pass raw or statistically filtered \texttt{ncu} metrics directly into the LLM's prompt and rely on the LLM itself to interpret them~\citep{zhang2025cudaforge,dong2026kernelblaster,li2026stitchcuda,nvidia2025kernelevolvemeta,zhang2026cudamaster}. cuPilot~\citep{chen2025cupilot} adds roofline-based bottleneck classification but stops at a label (memory-bound vs. compute-bound) without generating actionable directives from the underlying counters. In all cases, the LLM remains the sole interpreter of profiling data--no prior LLM-based system provides programmatic translation of hardware counters into optimization directives. The challenge predates LLMs: classical performance advisors such as GPA~\citep{zhou2021gpa} use instruction sampling and data-flow analysis to attribute stalls to their root causes and emit human-readable optimization suggestions, and independently observe that raw profiler output ``provide[s] little insight into how to improve the code.'' \model adopts the same diagnose-then-prescribe philosophy but \emph{closes the loop}: its micro-profiling tools translate counters into natural-language directives \emph{before} any LLM involvement, and an LLM then applies the fix directly rather than leaving it to a human engineer. Our ablation (Appendix~\ref{appendix:kernelbench-ablation}) confirms that this tool-based interpretation significantly outperforms both no-feedback and raw-metrics-only feedback.

\textbf{CUTLASS and CuTe code generation.} Several agents target NVIDIA's CUTLASS library, but at different abstraction levels. AVO~\citep{chen2026avo} and StitchCUDA~\citep{li2026stitchcuda} generate high-level CUTLASS template instantiations--selecting tile sizes, pipeline stages, and epilogue fusions from a predefined configuration space. \model operates at a lower level: it generates raw CUDA+CuTe source code by autonomously searching the CUTLASS/CuTe codebase for layout algebras, copy atoms, and MMA atoms, then composing them into novel kernels -- the same workflow used by expert kernel engineers.

\textbf{Tree search and evolutionary methods for LLM code optimization.} A growing body of work integrates Monte Carlo Tree Search with LLMs. MCTSr~\citep{zhang2024accessing} applies UCT to mathematical olympiad problems, where each node represents an answer version refined via self-evaluation. DeepSearch~\citep{wu2025deepsearch} embeds MCTS into the RL training loop with global frontier selection, targeting model weight updates rather than inference-time optimization. TreeRL~\citep{hou2025treerl} proposes entropy-guided token-level branching that forks from high-uncertainty tokens. Tree-GRPO~\citep{ji2025treesearch} uses complete agent steps as tree nodes and derives process supervision signals from intra-tree relative advantages. These systems treat tree search as a mechanism for \emph{training} LLMs or generating \emph{reasoning traces}. AlphaEvolve~\citep{novikov2025alphaevolve} takes an evolutionary approach, using population-based LLM sampling with scalar fitness to optimize algorithms including TPU kernels, but requires millions of samples over days of compute.

\model introduces a domain-adapted MCTS for LLM-based GPU kernel optimization, where each node represents a complete compiled-and-profiled CUDA kernel. This coarse granularity demands different design choices: progressive widening~\citep{auger2013continuous} gates expansion because each node requires compilation, execution, and profiling, ensuring exploration grows sublinearly with visit count; log-reward calibration eliminates the need for learned value functions; asymmetric branching factors reflect the empirical observation that valid optimizations outnumber valid repairs; dead-end pruning removes exhausted subtrees; and search memory enables cross-iteration learning. \model adopts MCTSr's ROOT re-expansion (fresh seed injection when existing subtrees plateau) but replaces self-reward with measured speedup, eliminating reward hacking risk. While writing this paper, we observed that OptiML~\citep{bhattacharjee2026optiml} has also independently applied MCTS with UCT for CUDA kernel optimization, which we view as convergent evidence that tree search is a natural fit for this domain.

\section{The Architecture of \model}

\subsection{High-Level Pipeline}

\model operates as a multi-agent system with three primary agents coordinated by a search orchestrator. The \textbf{Benchmarking Agent} (Stage 1) analyzes the reference implementation to establish baseline performance metrics and classify the kernel's bottleneck type (compute-bound, memory-bound, latency-bound, or mixed) via roofline analysis. The \textbf{Programmer Agent} generates optimized CUDA code in three modes: initial generation, debugging compilation/correctness errors, and iterative optimization based on profiling feedback. The \textbf{Profiling Agent} (Stage 2) executes micro-profiling tools filtered by the Stage 1 bottleneck classification, producing semantic feedback that guides the next optimization iteration. This two-stage architecture--bottleneck classification followed by filtered tool execution--reduces prompt noise and focuses the LLM on relevant optimizations.

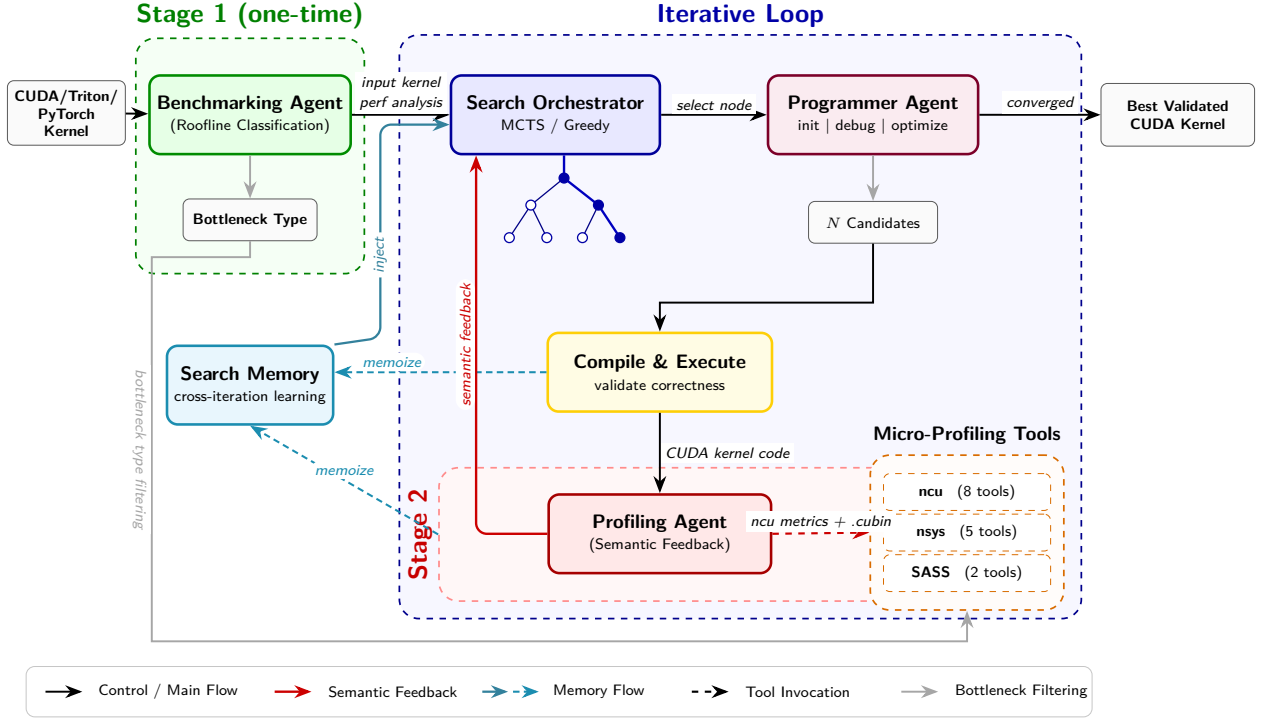
\begin{figure*}[t]
\centering
\resizebox{\textwidth}{!}{%
\begin{tikzpicture}[
    x=1cm, y=1cm,
    every node/.style={font=\sffamily\small},
    artifact/.style={rectangle, draw=gray!70!black, rounded corners=3pt, minimum height=0.68cm, align=center, fill=gray!4, font=\sffamily\scriptsize},
    toolbox/.style={rectangle, draw=orange!85!black, dashed, rounded corners=3pt, minimum width=2.72cm, minimum height=0.60cm, align=center, fill=white, font=\sffamily\scriptsize},
    agent/.style={rectangle, draw, very thick, rounded corners=5pt, minimum height=1.28cm, align=center, font=\sffamily\small},
    region/.style={rectangle, draw, dashed, thick, rounded corners=9pt},
    edgelabel/.style={fill=white, inner sep=2pt, font=\sffamily\itshape\scriptsize},
    mainflow/.style={-{Stealth[length=2.8mm,width=2.1mm]}, draw=black, line width=0.9pt},
    feedback/.style={-{Stealth[length=2.8mm,width=2.1mm]}, draw=red!80!black, line width=1.1pt, rounded corners=5pt},
    memoryflow/.style={-{Stealth[length=2.8mm,width=2.1mm]}, draw=cyan!55!black, line width=1.05pt, rounded corners=5pt},
    memoizeflow/.style={-{Stealth[length=2.8mm,width=2.1mm]}, draw=cyan!65!black, line width=0.95pt, densely dashed, rounded corners=5pt},
    toolinv/.style={-{Stealth[length=2.8mm,width=2.1mm]}, draw=red!80!black, line width=1pt, dashed},
    dataflow/.style={-{Stealth[length=2.8mm,width=2.1mm]}, draw=gray!70, line width=0.9pt},
    treeedge/.style={draw=blue!52!black, line width=0.6pt},
    treepath/.style={draw=blue!75!black, line width=1.1pt},
    treenode/.style={circle, draw=blue!65!black, fill=white, line width=0.6pt, minimum size=0.16cm, inner sep=0pt},
    selectednode/.style={treenode, fill=blue!65!black}
]
\node[region, draw=green!50!black, fill=green!4, minimum width=3.72cm, minimum height=3.83cm] (stageone) at (3.15,6.38) {};
\node[font=\sffamily\bfseries\large, text=green!50!black] at (3.15,8.67) {Stage 1 (one-time)};
\node[region, draw=blue!55!black, fill=blue!3, minimum width=11.10cm, minimum height=9.50cm] (loop) at (11.13,3.60) {};
\node[font=\sffamily\bfseries\large, text=blue!65!black] at (11.13,8.67) {Iterative Loop};
\node[region, draw=red!45, fill=red!3, minimum width=8.85cm, minimum height=2.17cm] (stagetwo) at (10.65,0.22) {};
\node[font=\sffamily\bfseries\large, text=red!78!black, rotate=90] at (5.92,0.22) {Stage 2};
\node[region, draw=orange!85!black, fill=orange!2, minimum width=3.15cm, minimum height=2.52cm] (toolpanel) at (14.82,0.25) {};
\node[font=\sffamily\bfseries\small] at (14.82,1.84) {Micro-Profiling Tools};
\node[artifact, minimum width=1.78cm, minimum height=1.05cm] (input) at (0.16,7.08) {\textbf{CUDA/Triton/}\\\textbf{PyTorch}\\\textbf{Kernel}};
\node[agent, draw=green!55!black, fill=green!10, minimum width=3.08cm] (benchmark) at (3.15,7.05) {\textbf{Benchmarking Agent}\\[-1pt]{\scriptsize (Roofline Classification)}};
\node[artifact, minimum width=2.18cm] (bottleneck) at (3.15,5.34) {\textbf{Bottleneck Type}};
\node[agent, draw=blue!60!black, fill=blue!9, minimum width=3.40cm] (search) at (8.12,7.05) {\textbf{Search Orchestrator}\\[-1pt]{\scriptsize MCTS / Greedy}};
\node[agent, draw=purple!65!black, fill=purple!8, minimum width=3.42cm] (programmer) at (13.29,7.05) {\textbf{Programmer Agent}\\[-1pt]{\scriptsize init $\mid$ debug $\mid$ optimize}};
\node[artifact, minimum width=2.10cm] (candidates) at (13.29,5.30) {$N$ Candidates};
\node[artifact, minimum width=2.50cm, minimum height=1.02cm] (output) at (18.25,7.05) {\textbf{Best Validated}\\\textbf{CUDA Kernel}};
\node[agent, draw=cyan!65!black, fill=cyan!8, minimum width=2.68cm] (memory) at (3.15,2.65) {\textbf{Search Memory}\\[-1pt]{\scriptsize cross-iteration learning}};
\node[agent, draw=yellow!65!orange, fill=yellow!13, minimum width=3.65cm] (execute) at (9.82,2.86) {\textbf{Compile \& Execute}\\[-1pt]{\scriptsize validate correctness}};
\node[agent, draw=red!65!black, fill=red!9, minimum width=3.62cm] (profiling) at (9.82,0.23) {\textbf{Profiling Agent}\\[-1pt]{\scriptsize (Semantic Feedback)}};
\node[toolbox] (ncu) at (14.82,0.91) {\textbf{ncu} \hspace{4pt}(8 tools)};
\node[toolbox] (nsys) at (14.82,0.25) {\textbf{nsys} \hspace{4pt}(5 tools)};
\node[toolbox] (sass) at (14.82,-0.41) {\textbf{SASS} \hspace{4pt}(2 tools)};
\draw[mainflow] (input) -- (benchmark);
\draw[mainflow] (benchmark) -- node[edgelabel, above, align=center] {input kernel\\[1pt]perf analysis} (search);
\draw[mainflow] (search) -- node[edgelabel, above] {select node} (programmer);
\draw[mainflow] (programmer) -- node[edgelabel, above] {converged} (output);
\draw[dataflow] (benchmark) -- (bottleneck);
\draw[dataflow] (programmer) -- (candidates);
\draw[mainflow] (candidates.south) -- ++(0,-0.96cm) -| (execute.north);
\draw[mainflow] (execute) -- node[edgelabel, right] {CUDA kernel code} (profiling);
\coordinate (filterturn) at ($(bottleneck.south)+(-1.60cm,-0.26cm)$);
\draw[dataflow] (bottleneck.south) -- ++(0,-0.26cm) -- (filterturn) -- node[midway, edgelabel, sloped, below, text=gray!70] {bottleneck type filtering} (1.55,-1.52) -- (14.82,-1.52) -- (toolpanel.south);
\coordinate (memoryentry) at ($(search.west)+(0,-0.16cm)$);
\coordinate (memorylane) at (5.27,3.42);
\draw[memoryflow] (memory.north east) -- (memorylane) -- node[midway, edgelabel, left, text=cyan!55!black, rotate=90] {inject} (memorylane |- memoryentry) -- (memoryentry);
\coordinate (compilememoryentry) at ($(memory.east)+(0,0.21cm)$);
\draw[memoizeflow] (execute.west) -- node[pos=0.73, edgelabel, above, text=cyan!65!black] {memoize} (compilememoryentry);
\draw[memoizeflow] (stagetwo.west) -- node[midway, edgelabel, above, text=cyan!65!black] {memoize} (memory.south);
\coordinate (feedbackentry) at ($(search.south west)+(0.43cm,0)$);
\draw[feedback] (profiling.west) -- (feedbackentry |- profiling.west) -- node[midway, edgelabel, sloped, above, text=red!80!black] {semantic feedback} (feedbackentry);
\draw[toolinv] (profiling.east) -- node[edgelabel, above] {ncu metrics + .cubin} (toolpanel.west);
\draw[treepath] ($(search.south)+(0.14cm,0)$) -- (8.26,6.02);
\draw[treeedge] (8.26,6.02) -- (7.72,5.58);
\draw[treepath] (8.26,6.02) -- (8.82,5.58) -- (9.17,5.05);
\draw[treeedge] (7.72,5.58) -- (7.37,5.05);
\draw[treeedge] (7.72,5.58) -- (7.98,5.05);
\draw[treeedge] (8.82,5.58) -- (8.56,5.05);
\node[selectednode] at (8.26,6.02) {};
\node[treenode] at (7.72,5.58) {};
\node[selectednode] at (8.82,5.58) {};
\node[treenode] at (7.37,5.05) {};
\node[treenode] at (7.98,5.05) {};
\node[treenode] at (8.56,5.05) {};
\node[selectednode] at (9.17,5.05) {};
\begin{scope}[yshift=-0.26cm]
\node[rectangle, draw=gray!45, rounded corners=4pt, fill=white, minimum width=17.35cm, minimum height=0.82cm] (legend) at (8.18,-2.08) {};
\draw[mainflow] (-0.18,-2.08) -- (0.42,-2.08);
\node[anchor=west, font=\sffamily\scriptsize] at (0.56,-2.08) {Control / Main Flow};
\draw[feedback] (3.55,-2.08) -- (4.15,-2.08);
\node[anchor=west, font=\sffamily\scriptsize] at (4.30,-2.08) {Semantic Feedback};
\draw[memoryflow] (6.93,-2.08) -- (7.35,-2.08);
\draw[memoizeflow] (7.43,-2.08) -- (7.85,-2.08);
\node[anchor=west, font=\sffamily\scriptsize] at (7.96,-2.08) {Memory Flow};
\draw[toolinv, draw=black] (10.34,-2.08) -- (10.94,-2.08);
\node[anchor=west, font=\sffamily\scriptsize] at (11.09,-2.08) {Tool Invocation};
\draw[dataflow] (13.75,-2.08) -- (14.35,-2.08);
\node[anchor=west, font=\sffamily\scriptsize] at (14.50,-2.08) {Bottleneck Filtering};
\end{scope}
\end{tikzpicture}%
}
\caption{
\model agentic optimization workflow. Stage~1 (Benchmarking Agent) performs one-time roofline-based bottleneck classification. The iterative loop comprises search, programming, compilation, correctness validation, profiling, and semantic feedback. The Search Orchestrator (MCTS or greedy) maintains an expansion tree over candidate solutions, while the Stage~2 Profiling Agent invokes bottleneck-filtered micro-profiling tools (8~\texttt{ncu} + 5~\texttt{nsys} + 2~SASS analyzers). Search Memory persists cross-iteration findings outside the loop. When the search converges, the workflow emits the best validated CUDA kernel.
}
\label{fig:pipeline}
\end{figure*}

All agents communicate with LLMs through a \textbf{backend-agnostic interface} built on the OpenAI Agents SDK~\citep{openai2025agentssdk}, so the same agent code runs unmodified on AWS Bedrock or vLLM~\citep{kwon2023vllm}. The Bedrock path requires the most adaptation: a custom adapter translates tool definitions, multi-turn tool-use/tool-result exchanges, and structured responses between the SDK's tool-calling protocol and the Bedrock Claude Messages API, whereas vLLM's OpenAI-compatible endpoint is consumed directly with no protocol translation.

\textbf{Proactive Tool Orchestration for the Profiling Agent.} A deliberate design choice in \model is the use of \emph{proactive} tool orchestration for the Profiling Agent, rather than the \emph{reactive} pattern typical of LLM agent frameworks~\citep{yao2023react}. In reactive systems, the LLM decides which tools to invoke--a stochastic process that may skip critical analyses. The Profiling Agent instead orchestrates its 15 micro-profiling tools \emph{deterministically}: all relevant tools are executed based on the bottleneck classification, injecting their guidance into the prompt before LLM invocation. This mirrors how human CUDA experts systematically analyze all relevant metrics rather than selectively querying information. The proactive approach guarantees comprehensive analysis and consistent optimization quality, which is essential for reliable CUDA optimization outcomes. A concrete example: in Task~41 (Appendix~\ref{appendix:case-l2-41}), the tools turn a memory-bottleneck diagnosis into a ranked list of specific, actionable directives (epilogue fusion, BF16 tensor cores, register-pressure and coalescing fixes); the agent applies the top recommendations together for a $2.8\times$ single-step improvement. A reactive agent invoking only a few tools would surface a fraction of this guidance.

\subsection{Agent Decomposition}

\subsubsection{Benchmarking Agent (Stage 1)}

The Benchmarking Agent runs \textbf{once per problem} at the start of optimization. It parses the reference code to identify the mathematical operations involved (GEMM, Softmax, convolution, etc.), computes a roofline estimate of theoretical peak performance from compute intensity and hardware specs, and classifies the problem as compute-bound, memory-bound, or mixed based on the CI/ridge ratio. It then profiles the reference implementation with \texttt{ncu} to validate this theoretical classification against measured utilization. The output is a \texttt{ReferenceAnalysis} object that guides downstream tool selection for the Programmer and Profiling Agents.

\subsubsection{Programmer Agent}

The Programmer Agent generates candidate kernel implementations through \model's backend-agnostic model interface, operating in three modes:
\begin{itemize}
    \item \textbf{Initial}: Generate first-attempt CUDA implementations from reference kernel
    \item \textbf{Debugging}: Fix compilation errors or correctness failures in previous attempts
    \item \textbf{Optimize}: Improve performance of working solutions based on profiler feedback
\end{itemize}

Prompts are constructed via a \textbf{modular PromptBuilder} that composes task-specific instructions from reusable components: base optimization strategies, dataset-specific formats, mode-specific instructions, and curated examples. This modularity allows different baseline types (PyTorch, Triton~\citep{tillet2019triton}, CUDA) to share common optimization knowledge while receiving tailored guidance. A notable feature is \textbf{GPU-aware prompt injection}: at agent creation time, \model detects the target GPU architecture and injects architecture-specific guidance---most importantly the correct tensor-core instruction shapes (\S\ref{sec:library-integration})---so generated code targets the right instruction format without requiring the LLM to memorize hardware details.

The agent outputs a structured \texttt{MultipleCompletions} object containing $N$ candidate solutions per iteration, each with its optimization strategy annotation. This structured output enables downstream validation and trajectory recording. A two-phase truncation-recovery mechanism handles cases where generated code exceeds the output token limit (detected via a \texttt{max\_tokens} stop reason): Phase A retries with doubled \texttt{max\_tokens}; if still truncated, Phase B retries at that doubled budget with an appended directive to restart concisely.

\subsubsection{Profiling Agent (Stage 2)}
\label{sec:profiling-agent}

The Profiling Agent transforms raw hardware metrics into semantic optimization guidance. Rather than simply forwarding \texttt{ncu} output to the LLM, it constructs a rich analysis prompt through a multi-stage process.

First, the agent collects metrics from three profiling sources: \textbf{\texttt{ncu}} provides kernel-level metrics (throughput, occupancy, stalls), \textbf{\texttt{nsys}} provides system-level timeline data (launch overhead, synchronization), and \textbf{SASS analysis} provides instruction-level information by disassembling the compiled \texttt{.so} file. To obtain the binary path, \model intercepts PyTorch's \texttt{load\_inline} compilation pipeline, capturing the ephemeral \texttt{.so} artifact for downstream \texttt{ncu} and SASS analysis. \texttt{ncu} profiling targets only the \model-generated \texttt{\_\_global\_\_} kernels identified via binary symbol analysis (\texttt{cuobjdump -symbols}, filtering for \texttt{STB\_GLOBAL} entry points), with fallback to duration-based kernel selection for library-call solutions (e.g., cuBLAS). This multi-level fusion is essential--\texttt{ncu} alone cannot distinguish whether a kernel uses Ampere-style async copies or Hopper TMA, but SASS reveals this directly.

Second, the agent invokes the \textbf{micro-profiling tool registry}. The key function \texttt{ToolRegistry.\allowbreak generate\_prompt\_guidance()} iterates through registered tools (filtered by Stage 1 bottleneck type), executes each tool's \texttt{analyze()} method on the collected metrics, and aggregates triggered findings into markdown-formatted guidance. This guidance is injected into the LLM prompt under an ``Automated Bottleneck Analysis'' section.

The resulting prompt structure follows a consistent format: SASS analysis results (tensor core instruction counts, memory patterns), followed by an nsys timeline summary (launch overhead percentage, sync calls)---where gap detection uses an adaptive threshold of $\max(1\%\ \text{of end-to-end time},\ 5\mu\text{s})$ so that only gaps meaningful relative to total runtime are flagged, avoiding false positives on long-running kernels---then tool-generated bottleneck analysis with severity ratings and recommendations, and finally ncu metrics organized by category (compute, memory, occupancy, stalls).

The LLM then synthesizes this information into a structured \texttt{ProfileAnalysis} output containing bottleneck classification, severity assessment, and prioritized optimization recommendations. This structured output feeds back to the Programmer Agent for the next iteration.

\textbf{Tool invocation pattern.} The Profiling Agent uses \emph{proactive} tool orchestration. With 15 micro-profiling tools, reactive invocation would introduce stochastic selection--the LLM might call 3 tools and skip 10, missing critical analyses. Instead, \model deterministically executes all relevant tools based on the Stage 1 bottleneck classification, injecting their combined guidance into the prompt \emph{before} LLM invocation. This guarantees comprehensive coverage and consistent optimization quality across runs. Stage 1 classifies the bottleneck using two independent methods for robustness: a theoretical roofline bound (arithmetic intensity vs.\ the hardware ridge point) and the measured compute-vs-memory throughput from \texttt{ncu}. Agreement between the two gives high confidence; on disagreement, \model defers to the theoretical bound, since it reflects the kernel's inherent ceiling rather than the current (possibly unoptimized) implementation. Table~\ref{tab:tool-filtering} shows the resulting tool filtering.

\begin{table*}[t]
\centering
\footnotesize
\caption{Tool filtering by bottleneck type}
\label{tab:tool-filtering}
\begin{tabular}{@{}lp{7.0cm}l@{}}
\toprule
\textbf{Bottleneck Type} & \textbf{Tool Selection} & \textbf{Rationale} \\
\midrule
COMPUTE\_BOUND & Exclude memory tools (Coalescing, DRAM) & Memory not limiting \\
MEMORY\_BOUND & Exclude compute tools (TensorCore, WGMMA) & Compute not limiting \\
LAUNCH\_OVERHEAD & Nsys tools (LaunchOverhead, SmallKernel) & System-level issues \\
MIXED/LATENCY\_BOUND & All tools & Multiple bottlenecks \\
\bottomrule
\end{tabular}
\end{table*}

\subsubsection{Open-Source Library Integration}
\label{sec:library-integration}

\model treats open-source GPU libraries as \textbf{first-class citizens} through an extensible integration architecture. Prior CUDA AI agents either stitch pre-compiled library calls (cuBLAS, cuDNN) or generate raw CUDA/Triton--none support source-level code generation with libraries like CUTLASS. \model's design enables LLM-based code generation with any GPU programming library through two mechanisms: (1) \textbf{GPU-aware prompt injection}--detecting target hardware and injecting library-specific configurations, and (2) a \textbf{library-aware build and runtime environment}--ensuring generated code compiles, executes, validates, and profiles seamlessly.

We demonstrate this architecture with \textbf{CUTLASS}, an industry-grade library for custom tensor-core kernels (adopted in systems such as PyTorch and TensorRT); its
\textbf{CuTe} layer provides the tensor and layout abstractions (copy
atoms, MMA atoms, layout algebra) used to compose such kernels from scratch. Generating
this code is challenging for LLMs: it must match hardware-specific instruction shapes and
satisfy architectural conventions that are easy to violate and sparsely documented,
and it depends on a precise build configuration.

For CUTLASS, \model's \textbf{prompt injection} detects the target GPU at runtime and injects architecture-specific instruction shapes, include paths, and compiler flags. \model's \textbf{build and runtime environment} handles compilation via PyTorch's \texttt{load\_inline} with CUTLASS-specific flags, tracks compilation events, validates correctness against reference implementations, and profiles compiled kernels with \texttt{ncu}/\texttt{nsys}. This closed-loop integration ensures that LLM-generated CUTLASS/CuTe code can compile, execute, profile, and evaluate seamlessly. Appendix~\ref{appendix:cutlass-case-studies} presents two complete trajectories demonstrating this integration end-to-end: a custom BF16 epilogue functor achieving $12.1\times$ (Task~41) and a dual-GEMM pipeline with hand-written kernels achieving $2.79\times$ (Task~46).

\label{sec:cutlass-search}
\textbf{LLM-autonomous code search.}
A key challenge in CUTLASS/CuTe code generation is that LLMs frequently produce code violating sparsely-documented architectural constraints (e.g., barrier lifecycle rules, warp specialization patterns, TMA pipeline depth). Rather than relying on static prompt injection of examples, \model equips the Programmer Agent with \textbf{LLM-autonomous code search tools} that allow it to query the CUTLASS/CuTe codebase directly during generation:

\begin{itemize}
    \item \texttt{search\_cutlass}: recursive grep over file contents with optional path/glob scoping (search calls are budgeted per iteration to encourage convergence toward generation; reads and listings are unbudgeted)
    \item \texttt{read\_cutlass\_file}: Read specific files with line ranges
    \item \texttt{list\_cutlass\_directory}: Directory listing for navigation
    \item \texttt{check\_smem\_layout}: pre-flight validation that compiles a minimal \texttt{tile\_to\_shape} probe with \texttt{nvcc} against the CUTLASS headers, catching shape/stride divisibility errors---the dominant CuTe compile failure---before a full kernel build (unbudgeted)
\end{itemize}

This design follows the principle that the LLM should \emph{discover} relevant patterns autonomously rather than receiving pre-selected examples. The LLM decides what to search based on the optimization context---e.g., retrieving \texttt{wgmma} pipeline patterns and MMA/copy atoms from the live CuTe source when generating Hopper tensor-core code---and composes them into novel kernels. This is the source-level workflow that expert engineers use, and the mechanism behind \model's from-scratch raw-CuTe kernels (Section~\ref{sec:veomni}). Contemporaneous work (GrepSeek;~\citealp{salemi2026grepseek}) independently finds that agent-issued executable search over a raw corpus can outperform index-based retrieval.

Additionally, a constraint mining tool pre-extracts hard architectural requirements from CUTLASS source (barrier lifecycle, warp specialization rules, vectorized epilogue patterns, tile swizzle) and injects them as non-negotiable code generation constraints. A companion tool validates generated code against these requirements before compilation, catching architectural errors early.

\subsubsection{Multi-Language Input Support}

\model supports \textbf{PyTorch-to-CUDA}, \textbf{Triton-to-CUDA}, and \textbf{CUDA-to-CUDA} translation. This flexibility serves diverse optimization workflows--from AI researchers prototyping in PyTorch to kernel engineers fine-tuning existing Triton or CUDA implementations.

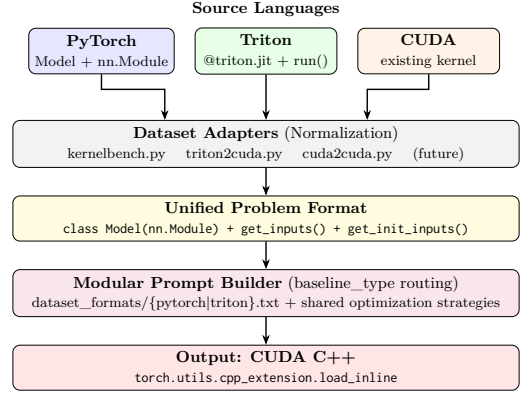
\begin{wrapfigure}{r}{0.44\textwidth}
\vspace{-1.5em}
\centering
\resizebox{0.96\linewidth}{!}{%
\begin{tikzpicture}[
    node distance=0.4cm,
    box/.style={rectangle, draw, rounded corners, minimum width=2.8cm, minimum height=0.9cm, align=center, font=\small},
    widebox/.style={rectangle, draw, rounded corners, minimum width=10cm, minimum height=0.8cm, align=center, font=\small},
    container/.style={rectangle, draw, dashed, rounded corners, inner sep=0.3cm},
    arrow/.style={-{Stealth[length=2mm]}, thick},
    label/.style={font=\footnotesize\itshape}
]

\node[box, fill=blue!10] (pytorch) {\textbf{PyTorch}\\{\footnotesize Model + nn.Module}};
\node[box, fill=green!10, right=0.4cm of pytorch] (triton) {\textbf{Triton}\\{\footnotesize @triton.jit + run()}};
\node[box, fill=orange!10, right=0.4cm of triton] (cuda) {\textbf{CUDA}\\{\footnotesize existing kernel}};

\node[above=0.1cm of triton, font=\small\bfseries] {Source Languages};

\node[widebox, fill=gray!10, below=0.9cm of triton] (adapters) {\textbf{Dataset Adapters} (Normalization)\\{\footnotesize kernelbench.py \quad triton2cuda.py \quad cuda2cuda.py \quad (future)}};

\node[widebox, fill=yellow!15, below=0.55cm of adapters] (unified) {\textbf{Unified Problem Format}\\{\footnotesize\texttt{class Model(nn.Module) + get\_inputs() + get\_init\_inputs()}}};

\node[widebox, fill=purple!10, below=0.55cm of unified] (prompt) {\textbf{Modular Prompt Builder} (baseline\_type routing)\\{\footnotesize dataset\_formats/\{pytorch|triton\}.txt + shared optimization strategies}};

\node[widebox, fill=red!10, below=0.55cm of prompt] (output) {\textbf{Output: CUDA C++}\\{\footnotesize\texttt{torch.utils.cpp\_extension.load\_inline}}};

\draw[arrow] (pytorch.south) -- ++(0,-0.3) -| ([xshift=-2cm]adapters.north);
\draw[arrow] (triton.south) -- (triton.south |- adapters.north);
\draw[arrow] (cuda.south) -- ++(0,-0.3) -| ([xshift=2cm]adapters.north);

\draw[arrow] (adapters) -- (unified);
\draw[arrow] (unified) -- (prompt);
\draw[arrow] (prompt) -- (output);

\useasboundingbox ([xshift=-2mm]adapters.west |- pytorch.north) rectangle
                  ([xshift=2mm]adapters.east |- output.south);

\end{tikzpicture}%
}
\caption{Multi-language input pipeline: all source languages are normalized to a unified format before being routed through language-specific prompts to produce optimized CUDA kernel.}
\label{fig:multi-lang-pipeline}
\vspace{-1em}
\end{wrapfigure}

While prior CUDA agents accept only PyTorch input (via KernelBench) or only existing CUDA code, \model natively supports PyTorch, Triton, and CUDA inputs---enabling optimization of kernels at any stage of the development pipeline (Figure~\ref{fig:multi-lang-pipeline}). All inputs are normalized to a unified \texttt{class Model} interface with language-specific prompt routing, producing CUDA C++ output via \texttt{load\_inline}.

This multi-language capability addresses a key production need: automating kernel optimization and hardware migration regardless of the source abstraction. Triton kernels hitting DSL-imposed performance ceilings (e.g., Ampere-era instructions on Hopper, Section~\ref{sec:veomni}), PyTorch prototypes requiring production CUDA implementations, and legacy CUDA kernels needing re-optimization for new GPU architectures can all be optimized through a single system---eliminating the need for scarce CUDA experts at each translation boundary.

\subsection{Search Strategy}

\model uses a solution-level \textbf{Monte Carlo Tree Search (MCTS)} strategy to navigate the CUDA optimization landscape. Each node is a complete compiled-and-profiled kernel rather than a single reasoning step. Unlike greedy search, MCTS systematically explores the multi-modal landscape (tiling vs vectorization vs algorithmic restructuring) while concentrating compute budget on the most promising subtrees. Algorithm~\ref{alg:mcts} presents the full MCTS loop.

\begin{algorithm*}[t]
\caption{MCTS Search with Progressive Widening and Search Memory. {\color{teal}Green lines} denote \model-specific adaptations for CUDA kernel optimization.}
\label{alg:mcts}
\begin{algorithmic}[1]
\State \textbf{Input:} Problem $P$, max iterations $T$, candidates per expansion $N$, UCT constant $C_{\text{uct}}$
\State $\text{ROOT} \gets \text{VirtualNode}()$, $\mathcal{M} \gets \emptyset$ \Comment{Virtual root, search memory}
\State $\text{ref\_analysis} \gets \text{BenchmarkingAgent.analyze}(P)$ \Comment{Stage 1: roofline classification}
\For{$t = 0$ to $T-1$}
    \State $\text{node} \gets \text{UCT-Select}(\text{ROOT}, C_{\text{uct}})$ \Comment{Traverse tree via UCT + progressive widening}
    \State {\color{teal}\textbf{if} node = ROOT \textbf{then} $\text{mode} \gets \text{initial}$; $\text{parent} \gets \text{None}$} \Comment{Fresh seed injection}
    \State {\color{teal}\textbf{else if} node.status = FAILED \textbf{then} $\text{mode} \gets \text{debug}$; $\text{parent} \gets \text{node}$} \Comment{Repair attempt}
    \State {\color{teal}\textbf{else} $\text{mode} \gets \text{optimize}$; $\text{parent} \gets \text{node}$} \Comment{Refine working solution}
    \State $\text{candidates} \gets \text{ProgrammerAgent.generate}(P, \text{parent}, \text{parent.feedback}, \mathcal{M}, \text{mode}, N)$
    \For{each $c$ in candidates}
        \State $\text{result} \gets \text{Execute}(c)$ \Comment{Compile, run, validate}
        \If{result.valid}
            \State {\color{teal}$\text{metrics} \gets \text{Profile}(c)$} \Comment{\texttt{ncu} + \texttt{nsys} hardware profiling}
            \State {\color{teal}$\text{sass\_info} \gets \text{SASSAnalyze}(c.\text{cubin})$} \Comment{Disassemble compiled binary}
            \State {\color{teal}$\text{guidance} \gets \text{RunTools}(\text{metrics}, \text{sass\_info}, \text{bottleneck})$} \Comment{Expert surrogates}
            \State {\color{teal}$\text{analysis} \gets \text{ProfilingAgent.analyze}(c, \text{metrics}, \text{guidance})$}
            \State {\color{teal}$c.\text{feedback} \gets \text{analysis}$} \Comment{Tool-guided feedback replaces rollout}
        \EndIf
        \State {\color{teal}$r(c) \gets \text{Reward}(c)$} \Comment{$\ln(\text{speedup})$, $-2.0$, or $-3.0$}
        \State $\text{Backpropagate}(c, r(c))$ \Comment{Update visits and rewards to ROOT}
        \State $\text{AttachChild}(\text{node}, c)$
    \EndFor
    \State {\color{teal}$\text{MarkTerminals}(\text{node})$} \Comment{Prune if $\geq$3 all-failed repair children}
    \State $\mathcal{M} \gets \text{ExtractMemory}(t, \text{candidates}, \text{result}, \mathcal{M})$ \Comment{Background summarizer}
\EndFor
\State \Return $\text{GlobalBest}(\text{ROOT})$ \Comment{Best solution across entire tree}
\end{algorithmic}
\end{algorithm*}

\subsubsection{UCT Selection}

\model's MCTS maintains a search tree where each node represents a concrete CUDA solution (compiled, profiled, and evaluated). Selection follows the UCT algorithm~\citep{kocsis2006bandit}: starting from a virtual ROOT node, the tree is traversed by repeatedly selecting the child that maximizes the upper confidence bound $\text{UCT}(c) = \bar{r}_c + C_{\text{uct}} \sqrt{\ln N_{\text{parent}} / N_c}$, where $\bar{r}_c$ is the node's mean reward, $N_c$ its visit count, $N_{\text{parent}}$ the parent's visit count, and $C_{\text{uct}} = \sqrt{2}$. Unlike evolutionary approaches that maintain a candidate population, tree-structured selection gives proper credit to underexplored subtrees--a node buried three levels deep in a promising branch competes fairly against shallow alternatives because the exploration bonus is computed relative to its parent's visits, not the global budget.

\subsubsection{Progressive Widening with Domain-Adapted Branching}
\label{sec:progressive-widening}

Standard MCTS assumes a finite action set and expands a node until all its actions are tried. \model's action--``generate another CUDA solution''--has an effectively unbounded branching factor, and each child is expensive to evaluate (compilation, execution, \texttt{ncu} profiling, and SASS analysis). \model therefore uses progressive widening~\citep{auger2013continuous} to gate expansion: a node with $N$ visits may have at most $C_{\text{pw}} \cdot N^{\alpha}$ children ($\alpha = 0.5$), so a node earns additional candidates only as repeated visits confirm its promise.

The key domain adaptation is asymmetric branching factors tied to node status. For ROOT (fresh seed generation) and FAILED nodes (repair attempts), the widening constant is $C_{\text{pw}} = 2$--moderate diversity is appropriate because failed solutions typically need a specific fix rather than broad exploration. For SUCCESSFUL nodes (optimization refinements), $C_{\text{pw}} = 3$--the search allocates more children because correct CUDA kernels have richer optimization surfaces with many orthogonal improvement axes (memory hierarchy tuning, instruction-level parallelism, tensor core utilization etc). This asymmetry encodes the empirical observation that the space of valid optimizations is broader than the space of valid repairs.

At each internal node during traversal, \model computes a \emph{virtual child UCT score}: the hypothetical UCT of a new child that would inherit its parent's mean reward as a prior and receive a fair share of visits. If this virtual score exceeds the UCT of the best existing child, the search expands the node (generates a new solution as its child) rather than descending further. This mechanism provides a principled expansion criterion--new children are only created when the uncertainty about unexplored alternatives outweighs the expected value of exploiting known children.

\subsubsection{Reward Calibration}

Backpropagation requires a scalar reward signal that meaningfully ranks solutions across the tree. \model uses a three-tier reward function calibrated on a diverse set of transformer-centric GPU kernels:

\begin{equation}
r(n) = \begin{cases}
\ln(\text{speedup}(n)) & \text{correct} \\
-2.0 & \text{incorrect output} \\
-3.0 & \text{crash/compile error}
\end{cases}
\end{equation}

The logarithmic transform for correct solutions prevents high-speedup outliers (e.g., 100$\times$ on trivially parallelizable kernels) from dominating the tree--a 2$\times$ speedup on a well-optimized kernel is arguably harder than 50$\times$ on an unoptimized one, and log-scaling reflects this. The gap between tiers (-2.0 for incorrect vs $\ln(1) = 0$ at the correctness boundary) ensures that ``almost correct'' solutions never dominate the tree over genuinely correct ones, preventing the search from wasting budget refining broken code. Rewards propagate from leaf to ROOT via backpropagation, updating visit counts and cumulative rewards at every ancestor.

\subsubsection{Dead-End Pruning and ROOT Re-expansion}

Two mechanisms prevent the search from stalling. First, \textbf{terminal marking}: when a failed node accumulates three or more repair children that all fail, it is marked terminal and excluded from future selection. This prunes dead-end subtrees where the underlying algorithmic approach is fundamentally broken, propagating upward through the tree as entire branches become exhausted.

Second, \textbf{ROOT re-expansion}: when UCT selects the virtual ROOT node itself (rather than any of its children), the system generates entirely fresh solutions from scratch rather than refining existing ones. This mechanism--inspired by MCTSr~\citep{zhang2024accessing}--injects exploration diversity as the tree deepens, preventing the search from over-committing to early seeds that happen to look promising. The progressive widening gate on ROOT ensures this happens at a controlled rate: fresh seeds are generated only when the virtual child's UCT score at ROOT exceeds that of all existing top-level solutions, indicating that unexplored regions of the solution space may outperform known optima.

A controlled ablation comparing MCTS against greedy search under matched compute budgets (Appendix~\ref{appendix:search-strategy-ablation}) confirms that tree-structured exploration yields a 26\% higher geometric mean speedup across 42 KernelBench tasks ($p = 0.004$), with the largest gains on problems requiring multi-step optimization chains where greedy converges prematurely. The NetVLAD trajectory (Appendix~\ref{appendix:cutlass-case-studies}, Task~46) provides concrete evidence: 43 consecutive failures precede the first working solution---a search depth that would exhaust any fixed-budget single-shot approach---yet once a correct kernel emerges, profiling-guided refinement converges rapidly ($0.75\times \to 2.79\times$ in two refinement steps).

\subsection{Search Memory: Cross-Iteration Learning}
\label{sec:session-memory}

A fundamental challenge in iterative LLM-based optimization is that each iteration starts \emph{tabula rasa}--the LLM has no memory of prior attempts beyond the immediate parent's profiling feedback. This leads to three failure modes we observe in practice: (1) \textbf{repeated errors}, where the LLM re-attempts approaches that failed in earlier iterations; (2) \textbf{lost discoveries}, where useful search findings (file paths, API patterns) from tool calls are not carried forward; and (3) \textbf{plateau behavior}, where the LLM cycles between similar strategies without building toward a solution.

\model addresses this with a \textbf{search memory} system (Figure~\ref{fig:pipeline})--a lightweight, append-only knowledge store scoped to a single optimization run. The system has three stages. First, a \textbf{post-iteration extractor}: after each iteration completes (success or failure), a separate background LLM call receives the iteration context--tool calls made, code approach attempted, and evaluation outcome--and produces structured entries for five memory sections. This separation ensures the code-generation LLM operates without meta-cognitive overhead. Second, a \textbf{structured memory store} organizes entries into semantically distinct sections: \texttt{Search Findings} (file paths, API patterns, and architectural details discovered via tool calls), \texttt{Errors \& Corrections} (what was tried and why it failed--compilation errors, correctness failures, performance regressions), \texttt{Successful Patterns} (approaches that achieved measurable speedup, with the technique and result), \texttt{Key Files} (source files containing relevant implementations or examples), and \texttt{Learnings} (higher-level principles distilled from the iteration, e.g., ``shared memory limits occupancy for this kernel size''). Third, \textbf{prompt injection}: the accumulated search memory is formatted as a structured block and injected into the Programmer Agent's system prompt at the start of each iteration. The memory grows monotonically across iterations, providing an increasingly rich context for optimization decisions.

A controlled ablation across 42 KernelBench tasks shows that search memory produces statistically equivalent final speedup while providing an early-convergence trend. Full ablation experiments are reported in Appendix~\ref{appendix:memory-ablation}.

\section{Micro-Profiling Tools}

\model's 15 micro-profiling tools encode GPU optimization expertise as composable analyzers that transform raw profiler metrics into actionable guidance. Each tool implements a trigger-analyze-recommend pattern: triggering on specific bottleneck signatures, analyzing severity and root cause, and generating targeted recommendations for the LLM. The tools are organized by profiler source (Table~\ref{tab:tool-categories})---kernel-level (\texttt{ncu}), instruction-level (SASS binary analysis), and system-level (\texttt{nsys})---because different bottlenecks manifest at different levels of abstraction and no single profiler provides complete visibility.

\begin{table*}[t]
\centering
\caption{Micro-profiling tool categories by profiler source}
\label{tab:tool-categories}
\begin{tabular}{@{}lcl@{}}
\toprule
\textbf{Category} & \textbf{Tools} & \textbf{Description} \\
\midrule
\texttt{ncu} (Nsight Compute) & 8 & Kernel-level metrics: throughput, stalls, occupancy \\
SASS (Binary Analysis) & 2 & Instruction-level: tensor cores, register spills \\
\texttt{nsys} (Nsight Systems) & 5 & System-level timeline: launch overhead, sync, transfers \\
\midrule
\textbf{Total} & \textbf{15} & \\
\bottomrule
\end{tabular}
\end{table*}

\subsection{Tools as Expert Surrogates}

As motivated in Section~\ref{sec:intro}, \model frames micro-profiling tools as \textbf{executable surrogates for human expert workflow}--encoding the profile-recognize-diagnose-prescribe cycle as callable tools. While demonstrated for CUDA, this pattern generalizes to any domain with measurable metrics, known bottleneck patterns, and actionable fixes (e.g., database query optimization, compiler tuning).

\subsubsection{The 5-Tuple Formalization}

Each \model micro-profiling tool encodes one expert heuristic as a formal 5-tuple:

\begin{equation}
    \tau = (\text{name}, \mathcal{M}_{\text{req}}, \theta, \text{trigger}, \text{analyze})
\end{equation}

\noindent where $\mathcal{M}_{\text{req}} \subseteq \mathcal{M}$ is the required metric subset (from the ${\sim}50$ \texttt{ncu} metrics \model collects), $\theta$ are threshold parameters, $\text{trigger}: \mathcal{M} \to \{0, 1\}$ is the activation predicate, and $\text{analyze}: \mathcal{M} \to \mathcal{O}$ produces structured output guidance. The output space $\mathcal{O}$ is a fixed schema comprising: a \emph{severity} level (critical/high/medium/low), a \emph{root cause} explanation of why the bottleneck exists, ranked \emph{recommendations} with expected improvement estimates, and optionally a \emph{code example} demonstrating the fix. This schema is what distinguishes semantic feedback from raw metric forwarding--the LLM receives not ``occupancy = 6\%'' but ``occupancy is critically low (6\%) because shared memory usage limits concurrent blocks to 2; switch to warp-level reduction for an expected 3--5$\times$ improvement.'' A full worked example is provided in Appendix~\ref{appendix:tool-specs}. The following table illustrates how the 5-tuple maps to expert reasoning:

\begin{table*}[t]
\centering
\caption{Mapping human expert workflow to tool components}
\begin{tabular}{@{}ll@{}}
\toprule
\textbf{Expert Thought Process} & \textbf{Tool Component} \\
\midrule
\multicolumn{2}{@{}l@{}}{\textit{Example 1: TensorCoreUnderutilizationTool}} \\
``Let me check tensor core usage'' & $\mathcal{M}_{\text{req}} = \{\texttt{TC\_util}\}$ \\
``If it's below 10\%, that's critical'' & $\theta = \{\text{low}: 10\%\}$ \\
``This looks like matmul without TC'' & $\text{trigger}(m) = \mathds{1}[\texttt{TC\_util} < 10\%]$ \\
``They should use cuBLAS'' & $\text{analyze}() \to \text{recommendations}$ \\
\midrule
\multicolumn{2}{@{}l@{}}{\textit{Example 2: WarpStallTool}} \\
``Why are warps not issuing?'' & $\mathcal{M}_{\text{req}} = \{\texttt{mem\_stall}, \texttt{barrier\_stall}, ...\}$ \\
``Over 40\% stall on one reason is bad'' & $\theta = \{\text{high}: 40\%\}$ \\
``Memory dependency is the bottleneck'' & $\text{trigger}(m) = \mathds{1}[\max(\text{stalls}) > 40\%]$ \\
``Use shared memory to hide latency'' & $\text{analyze}() \to \text{stall-specific fixes}$ \\
\bottomrule
\end{tabular}
\end{table*}

\subsubsection{Proactive vs. Reactive Tool Orchestration}

\model uses \textbf{proactive} tool orchestration for the Profiling Agent: rather than letting the LLM decide which tools to call (the reactive pattern in standard agent frameworks), \model deterministically executes all bottleneck-relevant tools and injects their combined guidance into the prompt before LLM invocation. Crucially, ``bottleneck-relevant'' is not ``all tools''--the bottleneck classification from Stage~1 feeds a tool-affinity filter that excludes irrelevant analyses (e.g., memory-bound kernels skip tensor core utilization checks, compute-bound kernels skip coalescing analysis), preventing signal dilution while guaranteeing comprehensive coverage of the actual bottleneck. This transforms expert knowledge from an optional resource into a guaranteed input. The reactive pattern's weakness--stochastic tool selection leading to missed analyses and inconsistent coverage--is particularly harmful for CUDA optimization, where the relevant bottleneck is often non-obvious and experts systematically check all metrics rather than guessing which to examine. Our controlled ablation (Appendix~\ref{appendix:tool-mode-ablation}) confirms this empirically: proactive orchestration achieves 23\% higher geometric mean speedup than reactive function calling ($p = 0.035$).

\subsubsection{Pluggability and User Extensibility}

\model's tools are \textbf{pluggable}--they can be dynamically added, removed, or modified without changing the core system via a decorator-based registration pattern (\texttt{@ToolRegistry.register}). Users subclass \texttt{ProfilingTool}, declare required metrics and thresholds, and implement an \texttt{analyze()} method; the tool is then automatically incorporated into the proactive orchestration pipeline. This makes \model a platform rather than a fixed tool: teams can encode domain-specific heuristics (e.g., MoE expert load imbalance detection), add support for new GPU architectures before official updates, or register proprietary optimizations as private tools. The full registration API and a worked example are provided in Appendix~\ref{appendix:tool-specs} (Listing~\ref{lst:tool-reg}).

\subsection{Tool Specifications}

Each tool follows the 5-tuple formalization described above. The complete tool taxonomy (15 tools organized by category and bottleneck type) and detailed specifications are provided in Appendix~\ref{appendix:tool-specs}, including full reference implementations (Listings~\ref{lst:tool-reg} and~\ref{lst:warpstall}) that make the trigger-analyze-recommend pattern concrete. We detail two representative tools below: one NCU-based analyzer and the SASS binary analysis tools.

\subsubsection{Warp Stall Analyzers}

\model includes two warp stall analyzers built on \texttt{ncu} metrics. The primary \texttt{WarpStallTool} analyzes why warps are not issuing instructions, identifying the dominant stall reason. It requires five warp-stall metrics $\mathcal{M}_{\text{req}} = \{\texttt{mem\_dep},\; \texttt{short\_scoreboard},\; \texttt{long\_scoreboard},\; \texttt{barrier},\; \texttt{branch\_resolving}\}$, where $m.s$ denotes the percentage of active cycles stalled on reason $s$. The tool finds the dominant stall $s^* = \arg\max_{s \in \mathcal{M}_{\text{req}}} m.s$ and triggers when $m.s^*$ exceeds $\theta_{\text{high}} = 40\%$ (severity escalates to critical at $60\%$). A companion \texttt{BarrierStallTool} triggers at a lower threshold ($30\%$) specifically for barrier synchronization issues, which are both a performance and correctness concern.

\begin{align*}
    \mathcal{M}_{\text{req}} &= \bigl\{\,\texttt{mem\_dep},\; \texttt{short\_scoreboard},\; \texttt{long\_scoreboard},\\
    &\qquad\;\; \texttt{barrier},\; \texttt{branch\_resolving}\,\bigr\} \\[4pt]
    s^* &= \arg\max_{s \in \mathcal{M}_{\text{req}}} m.s \\[4pt]
    \theta &= \{\text{high}: 40\%,\; \text{critical}: 60\%\} \\[4pt]
    \text{trigger}(m) &= \mathds{1}\bigl[m.s^* > \theta_{\text{high}}\bigr] \\[4pt]
    \text{analyze}(m) &\rightarrow \bigl(\text{severity by } m.s^*;\ \text{root cause and fix specific to } s^*\bigr)
\end{align*}

When triggered, the tool translates the raw stall percentage into a structured diagnostic: a \emph{root cause} explaining why the stall occurs (e.g., ``warps are waiting for global memory operations to complete---400+ cycle latency with no prefetching''), followed by ranked \emph{actionable recommendations} tailored to the dominant stall type. Memory dependency stalls produce guidance such as ``use shared memory to cache frequently accessed data'' and ``implement vectorized loads (\texttt{float4}) for better bandwidth''; barrier stalls recommend ``use warp-level primitives (\texttt{\_\_shfl\_sync}, \texttt{\_\_ballot\_sync}) instead of \texttt{\_\_syncthreads()}'' with a code example showing the correct pattern; scoreboard stalls suggest increasing instruction-level parallelism or using async copies (\texttt{cp.async}) on Ampere+; and branch resolving stalls recommend replacing conditionals with arithmetic predication (e.g., \texttt{result = cond * val1 + (1-cond) * val2}). This translation is what enables the LLM to act on profiler output without requiring GPU architecture expertise---the tool bridges the gap between ``47\% memory dependency stalls'' (a number) and ``add shared memory tiling with \texttt{\_\_syncthreads()} between load and compute phases'' (an action). The complete implementation, including the per-stall-type root causes and recommendations, is given in Listing~\ref{lst:warpstall} (Appendix~\ref{appendix:tool-specs}).

\subsubsection{SASS Binary Analysis Tools}
\label{sec:sass-tools}

While \texttt{ncu} tools analyze \emph{runtime counters}, SASS tools analyze the \emph{compiled binary itself} (what hardware instructions were actually emitted by \texttt{nvcc}). \texttt{ncu} can report ``tensor core utilization = 0\%'' but cannot explain whether the kernel lacks tensor core instructions entirely, uses an older generation (HMMA vs WGMMA), or delegates to cuBLAS (whose tensor-core kernels run as separate library launches, absent from the user kernel's compiled \texttt{.so}). \model extracts SASS via \texttt{cuobjdump --dump-sass} on the compiled \texttt{.so} and counts instruction patterns across five tensor core generations (WGMMA, HMMA, IMMA, DMMA, BMMA), memory operations (LDGSTS, TMA), and register spilling (STL/LDL). Full 5-tuple specifications for both SASS tools are given in Appendix~\ref{appendix:tool-specs}.

\textbf{WGMMAInstructionTool} provides definitive hardware feature detection. When no tensor core instructions appear in SASS, it checks for cuBLAS dynamic symbols via \texttt{nm -D}---distinguishing ``no TC instructions because the kernel is element-wise'' from ``no TC instructions because cuBLAS handles them at runtime.'' On Hopper targets, if HMMA instructions are found but no WGMMA, it recommends switching to Hopper warpgroup MMA (expected 1.3--1.5$\times$ improvement). In the VeOmni case study (Section~\ref{sec:veomni}) this mechanism flagged 37 candidates that compiled but emitted zero tensor-core instructions and an \texttt{HMMA}-with-register-spill regression, steering the search toward the winning WGMMA kernel.

\textbf{RegisterSpillDetector} combines SASS evidence (STL/LDL counts prove spills exist) with \texttt{ncu} metrics to determine whether spills \emph{actually} cause harm. Not all spills are pathological, so the tool triggers only under three multi-signal conditions---spill-induced latency bubbles, bandwidth contention from local-memory traffic, or register-bound occupancy that starves the scheduler---avoiding false alarms on harmless spills while catching cases where register pressure is the root bottleneck. (Exact thresholds in Appendix~\ref{appendix:tool-specs}.)

\section{Results and Evaluation}

We evaluate \model along three axes---SOTA comparison on KernelBench, controlled ablations of each design component, and production validation on expert-optimized kernels---and close with a preliminary energy-aware extension. Throughout all experiments, a kernel is \emph{numerically correct} if its output satisfies \texttt{torch.allclose(out, ref, atol=1e-2, rtol=1e-3)}.

\subsection{Experimental Setup}

\textbf{Benchmark.} KernelBench~\citep{ouyang2024kernelbench} comprises 250 GPU kernel tasks at three difficulty levels: Level~1 (100 single operators), Level~2 (100 fused/composed operators), and Level~3 (50 full model architectures). Speedup is measured as wall-clock time of the PyTorch eager baseline divided by the optimized CUDA kernel.

\textbf{Configuration.} Unless otherwise noted, experiments use Claude Sonnet 4.6 on NVIDIA A100 GPUs (AWS p4 instances), with 15 seeds per task (5 temperatures $\times$ 3 rounds), 30 iterations, 2 candidates per iteration, and MCTS search. A full comparison with concurrent systems (hardware, metric type, task subsets) is provided in Appendix~\ref{appendix:comparison}.

\subsection{State-of-the-Art Comparison on KernelBench}

\begin{table*}[t]
\centering
\caption{\model vs.\ KernelBlaster on KernelBench (geometric mean speedup, A100). KernelBlaster was the prior SOTA at time of submission.}
\label{tab:kernelblaster-comparison}
\begin{tabular}{@{}lcccc@{}}
\toprule
\textbf{System} & \textbf{Level 1} & \textbf{Level 2} & \textbf{Level 3} & \textbf{Solved} \\
\midrule
\textbf{\model} & \textbf{2.42$\times$} & \textbf{4.69$\times$} & \textbf{5.30$\times$} & 100/100/50 \\
KernelBlaster & 1.43$\times$ & 2.50$\times$ & 1.50$\times$ & -- \\
\midrule
\textbf{\model improvement} & +69\% & +88\% & +253\% & \\
\bottomrule
\end{tabular}
\end{table*}

\model outperforms KernelBlaster on all three levels, with the largest margin on Level~3 (+253\%)---full model architectures where MCTS-guided multi-step optimization and micro-profiling tools provide the most actionable guidance. \model achieves complete task coverage: 100/100 on Level~1, 100/100 on Level~2, and 50/50 on Level~3. Despite KernelBlaster's use of RL-style exploration and a persistent knowledge base, \model achieves higher speedups with a simpler approach grounded in micro-profiling tools as expert surrogates. A comprehensive comparison with five additional concurrent systems (CudaForge, StitchCUDA, KernelFoundry, AVO, and others) confirming \model's advantage across hardware platforms and metric types is provided in Appendix~\ref{appendix:comparison}.

\textbf{Robustness to outliers.} Since some Level~2 and Level~3 kernels have trivially parallelizable reference implementations (yielding speedups $>$100$\times$), we verify that \model's advantage is not driven by a few extreme outliers. Table~\ref{tab:capped-geomean} reports geometric mean speedups under progressively aggressive capping thresholds.

\begin{table}[t]
\centering
\small
\caption{Geometric mean speedup under capping thresholds. Each task's speedup is clamped to $\min(\text{speedup}, \text{cap})$ before computing the geometric mean. Results demonstrate broad-based gains across all levels.}
\label{tab:capped-geomean}
\begin{tabular}{@{}lccccccc@{}}
\toprule
\textbf{Level} & \textbf{Solved} & \textbf{No cap} & \textbf{Cap@100} & \textbf{Cap@50} & \textbf{Cap@20} & \textbf{Cap@10} & \textbf{Cap@5} \\
\midrule
L1 & 100/100 & 2.42$\times$ & 2.42$\times$ & 2.42$\times$ & 2.42$\times$ & 2.34$\times$ & 2.09$\times$ \\
L2 & 100/100 & 4.69$\times$ & 4.26$\times$ & 3.99$\times$ & 3.62$\times$ & 3.25$\times$ & 2.61$\times$ \\
L3 & 50/50 & 5.30$\times$ & 5.08$\times$ & 4.95$\times$ & 4.66$\times$ & 4.31$\times$ & 3.57$\times$ \\
\bottomrule
\end{tabular}
\end{table}

Level~1 speedups show minimal sensitivity to capping (maximum task speedup is 21.4$\times$; geometric mean is unchanged through Cap@20), confirming uniformly distributed gains. Level~3 retains 4.31$\times$ even at Cap@10, demonstrating that its strong performance is broad-based rather than driven by outliers. Level~2 shows the largest sensitivity to capping (4.69$\times$ $\to$ 3.25$\times$ at Cap@10) due to several reference implementations with extreme inefficiency; even under Cap@5, \model still achieves 2.61$\times$---well above all competing systems.

\subsection{Component Ablation}
\label{sec:component-ablation}

We isolate the contribution of each \model design component through controlled ablations on 42 KernelBench tasks (L1:~19, L2:~16, L3:~7) selected as a representative subset of transformer-relevant workloads, enabling controlled ablation at scale across multiple conditions, seeds, and iterations that would be prohibitive on the full 250-task benchmark. Each ablation varies a single component while holding all others fixed. All ablations use the same 42-task set; the effective sample size $n$ reported in Wilcoxon tests may be smaller because the test excludes tied pairs and tasks where a condition fails to produce a valid kernel. Table~\ref{tab:ablation-summary} summarizes the results; full experimental details are in the referenced appendices.

\begin{table*}[t]
\centering
\small
\caption{Consolidated ablation: independent contribution of each \model component. Effect size is the ratio of geometric mean speedup (treatment/control). All use Wilcoxon signed-rank test (one-sided).}
\label{tab:ablation-summary}
\begin{tabular}{@{}lllccl@{}}
\toprule
\textbf{Component} & \textbf{Treatment} & \textbf{Control} & \textbf{Geo Mean} & \textbf{$p$-value} & \textbf{Details} \\
\midrule
Micro-profiling tools & Full pipeline & Raw \texttt{ncu} only & 4.00$\times$ vs 1.77$\times$ (+125\%) & $<$0.0001*** & App.~\ref{appendix:kernelbench-ablation} \\
MCTS search & Tree search & Greedy & 4.60$\times$ vs 3.65$\times$ (+26\%) & 0.004** & App.~\ref{appendix:search-strategy-ablation} \\
Proactive orchestration & Deterministic & Reactive & 1.37$\times$ vs 1.12$\times$ (+23\%) & 0.035* & App.~\ref{appendix:tool-mode-ablation} \\
Search memory & Memory ON & Memory OFF & 1.82$\times$ vs 1.72$\times$ (+6\%) & 0.181 (n.s.) & App.~\ref{appendix:memory-ablation} \\
\bottomrule
\end{tabular}
\end{table*}

Three components independently and significantly improve optimization quality ($p < 0.05$). We discuss the most informative findings from each ablation below; MCTS provides the second-largest effect (+26\%) by enabling escape from local optima that trap greedy search. Proactive orchestration (+23\%) ensures comprehensive tool coverage versus stochastic reactive invocation. Search memory shows a positive but non-significant trend on final speedup (+6\%, $p = 0.181$), while improving early convergence.

\subsubsection{Raw Metrics Are Harmful}

The tool ablation reveals a counterintuitive result: providing raw \texttt{ncu} metrics without interpretation performs significantly \emph{worse} than providing no profiling feedback at all (geometric mean 1.77$\times$ vs 3.35$\times$, Wilcoxon $p = 0.0007$). This condition passes ${\sim}$50 raw hardware counters as name-value pairs---comparable to the approach used by CudaForge~\citep{zhang2025cudaforge}. Unstructured metric dumps distract the LLM: it attempts to optimize counters that do not translate to wall-clock improvements, or misinterprets raw values without roofline context. \model's full pipeline (4.00$\times$) outperforms both controls, confirming that \emph{interpretation}---not data access---drives optimization quality. The Task~41 trajectory (Appendix~\ref{appendix:case-l2-41}) illustrates this concretely: the tools translate the raw counters (15.88 sectors/request, 8.7\,GB read traffic, 244 registers/thread) into concrete directives---fuse the epilogue, switch to BF16 tensor cores---that produce a $2.8\times$ single-step improvement; the same counters presented unstructured carry no such actionable directive. This finding generalizes beyond KernelBench: on 6 representative Triton inference kernels (Appendix~\ref{app:ablation-production}), tools achieve 36\% higher speedup in 31\% fewer iterations.

\begin{table}[t]
\centering
\small
\caption{Tool ablation: correctness and speedup across 42 KernelBench tasks. Unsolved tasks scored as 1.0$\times$.}
\label{tab:ablation-main}
\begin{tabular}{@{}lccc@{}}
\toprule
\textbf{Configuration} & \textbf{Solved} & \textbf{Geo Mean} & \textbf{Win Rate} \\
\midrule
No Feedback & 32/42 (76\%) & 3.35$\times$ & 10/42 (24\%) \\
Raw \texttt{ncu} Metrics & 36/42 (86\%) & 1.77$\times$ & 3/42 (7\%) \\
\textbf{\model (full pipeline)} & \textbf{42/42 (100\%)} & \textbf{4.00$\times$} & \textbf{25/42 (60\%)} \\
\bottomrule
\end{tabular}
\end{table}

Table~\ref{tab:ablation-wilcoxon-main} reports pairwise Wilcoxon signed-rank tests confirming the statistical ordering Ours $>$ No Feedback $>$ Raw \texttt{ncu}. Three insights emerge: (1) the interpretation gap (Raw $\to$ Ours) is larger than the data gap (None $\to$ Ours), confirming that \emph{how} profiling data is presented matters more than \emph{whether} it is present; (2) Raw \texttt{ncu} is significantly \emph{worse} than No Feedback ($p = 0.0007$), meaning unstructured metrics actively degrade optimization quality---a direct challenge to the design of systems like CudaForge that pass raw counters; (3) $n = 39$--$40$ out of 42 tasks show non-zero differences, indicating the effect is broad-based rather than driven by a few outlier tasks.

\begin{table}[t]
\centering
\small
\caption{Pairwise Wilcoxon signed-rank tests (one-sided, $H_1$: first $>$ second).}
\label{tab:ablation-wilcoxon-main}
\begin{tabular}{@{}lccc@{}}
\toprule
\textbf{Comparison} & $n$ & $W$ & $p$ \\
\midrule
Ours $>$ Raw \texttt{ncu} & 40 & 728 & $<$0.0001*** \\
Ours $>$ No Feedback & 40 & 590 & 0.0078** \\
No Fdbk $>$ Raw \texttt{ncu} & 39 & 618 & 0.0007*** \\
\bottomrule
\end{tabular}
\end{table}

\subsubsection{Which Tools Drive the Largest Gains?}

Per-tool impact analysis across 42 tasks reveals which tools most reliably translate into optimization gains. Table~\ref{tab:tool-impact-main} summarizes tool coverage and effectiveness; the \emph{hit rate}---fraction of fires producing $>$1.5$\times$ improvement---measures how reliably the LLM acts on each tool's guidance.

\begin{table}[t]
\centering
\small
\caption{Per-tool impact across 42 KernelBench tasks. ``Fires'' counts optimization turns where the tool's recommendation appears; ``Hit Rate'' is the fraction producing $>$1.5$\times$ improvement.}
\label{tab:tool-impact-main}
\begin{tabular}{@{}lcccc@{}}
\toprule
\textbf{Tool} & \textbf{Tasks} & \textbf{Fires} & \textbf{Hit Rate} & \textbf{Avg.} \\
\midrule
RegisterSpill & 16/42 & 44 & \textbf{18.2\%} & 1.44$\times$ \\
OccupancyLimiter & 35/42 & 241 & 13.3\% & 1.48$\times$ \\
WarpStall & 35/42 & 318 & 11.0\% & 1.41$\times$ \\
SharedMemTiling & 28/42 & 172 & 9.3\% & 1.42$\times$ \\
LaunchOverhead & 24/42 & 141 & 8.5\% & 1.32$\times$ \\
TensorCore/cuBLAS & 42/42 & 835 & 8.1\% & 1.31$\times$ \\
MemoryCoalescing & 37/42 & 439 & 5.9\% & 1.24$\times$ \\
Vectorization & 41/42 & 595 & 4.0\% & 1.16$\times$ \\
\bottomrule
\end{tabular}
\end{table}

The tensor core switch is the single highest-impact action: for compute-bound kernels where the LLM generates CUDA-core-only code, the TensorCoreUnderutilizationTool's recommendation to use cuBLAS/CUTLASS produces an average \textbf{6.25$\times$} gain across 9 tasks (max 9.6$\times$)---Task~41's $4.11\times \to 11.49\times$ single-iteration leap to a fused BF16 CUTLASS GEMM (Appendix~\ref{appendix:case-l2-41}) exemplifies this pattern. RegisterSpillDetector has the highest per-fire reliability (18.2\%), because its diagnostic (``168 registers/thread, spilling to local memory'') is immediately actionable---the LLM can directly restructure loops to reduce live variables. Memory optimization tools (coalescing, vectorization) fire most broadly (37--41 of 42 tasks) but convert at lower rates (4--6\%), reflecting the difficulty of non-trivial data layout changes.

\subsubsection{MCTS Escapes Local Optima}

The 26\% aggregate improvement from MCTS masks dramatic per-task effects. On tasks with multi-modal optimization landscapes, MCTS achieves up to \textbf{10.0$\times$} the speedup of greedy search on the same task (L1 task~7: MCTS 25.0$\times$ vs greedy 2.5$\times$; L2 task~33: MCTS 8.0$\times$ vs greedy 1.0$\times$; L3 task~31: MCTS 25.0$\times$ vs greedy 4.2$\times$). In these cases, greedy converges to a correct but suboptimal solution and cannot escape its local basin, while MCTS explores alternative subtrees that discover fundamentally superior optimization paths (e.g., switching from shared-memory tiling to warp-shuffle reduction). Greedy wins on 8 tasks---typically unimodal landscapes where depth-first exploitation suffices and MCTS wastes budget on unnecessary exploration.

\subsection{Production Validation: VeOmni MoE}
\label{sec:veomni}

We further apply \model to \textbf{VeOmni}, a production MoE training
stack whose grouped-GEMM kernels are hand-optimized in Triton by expert engineers. We target the
\texttt{dW1} weight-gradient kernel (\texttt{group\_gemm\_same\_mn}) from
GPT-OSS-120B's \texttt{gate\_up\_proj} backward pass: per expert $e$ it computes
$\mathrm{dW1}_e = X_e^{\top} G_e$ ($M{=}2880$, $N{=}5760$, $E{=}128$), with a
\emph{ragged} contraction $K_e$ that varies per expert under top-4 routing (total
$R = \sum_e K_e = 57{,}768$ rows; per-expert load from a measured-routing imbalance
model, Appendix~\ref{appendix:veomni-dw1-source}). This is harder than a forward
GEMM---the imbalance falls on the \emph{reduction} axis, so each expert accumulates
over a variable-length $K_e$ while writing a fixed $[M,N]$ tile.

On a single H100, \model achieved a \textbf{$1.23\times$ speedup} over the expert-tuned
Triton baseline; beyond this best kernel, it found \textbf{16 other distinct correct
kernels} that also beat the baseline. Table~\ref{tab:veomni-dw1} traces the MCTS search
climbing from a $0.07\times$ first attempt to $1.23\times$ over 18 iterations, each leap
driven by the bottleneck \model's profiling stage surfaced at the prior node. The gain
comes not from tensor cores per se---the Triton baseline already issues Hopper warp-group
MMA---but from shape-specific tuning of the memory path and occupancy for this
transposed, ragged-$K$ layout.

\begin{table}[t]
\centering
\caption{\model search progression on the VeOmni \texttt{dW1} weight-gradient kernel
(H100). Bottlenecks are those flagged by \model's Stage-2 profiling at each milestone.}
\label{tab:veomni-dw1}
\begin{tabular}{@{}clc@{}}
\toprule
\textbf{Iter} & \textbf{Dominant bottleneck addressed} & \textbf{Speedup} \\
\midrule
--   & Triton baseline (\texttt{group\_gemm\_same\_mn}) & $1.00\times$ \\
0    & Initial raw-CuTe WGMMA (SMEM bank conflicts)     & $0.07\times$ \\
8    & Uncoalesced global loads (cp.async staging)      & $0.78\times$ \\
14   & Uncoalesced loads (swizzled SMEM)                & $0.97\times$ \\
16   & Uncoalesced loads (vectorized epilogue)          & $1.15\times$ \\
\textbf{18} & \textbf{Occupancy / 168 regs/thread ($128{\times}128$ tile, 4-stage)} & $\mathbf{1.23\times}$ \\
\bottomrule
\end{tabular}
\end{table}

\paragraph{Native CuTe\,+\,raw-CUDA generation.}
This kernel is the clearest production demonstration of \model's headline claim
(Contribution~4): autonomous \emph{source-level} CuTe code generation, not high-level
template instantiation or library stitching. The winning solution contains
\textbf{zero library calls}---no cuBLAS, no CUTLASS template---and
is instead a from-scratch Hopper kernel that \model composed by searching the
CUTLASS/CuTe codebase during generation. At the source level it assembles: an
\texttt{SM90\_64x128x16} \textbf{WGMMA} warp-group MMA atom driven by
\texttt{cute::gemm} with explicit \texttt{warpgroup\_arrive}/\texttt{commit\_batch}/%
\texttt{wait} fences; a \textbf{four-stage \texttt{cp.async} software pipeline}
prefetching tiles into shared memory; \textbf{128-byte-swizzled shared memory}
(\texttt{GMMA::Layout\_MN\_SW128\_Atom}) for bank-conflict-free WGMMA operand reads;
and \textbf{CuTe layout-algebra predication} (\texttt{make\_identity\_tensor} with
\texttt{copy\_if}) to mask the ragged per-expert $K_e$ boundaries. Notably, \model
reasoned explicitly that a single TMA descriptor would mis-read neighboring experts'
rows under ragged $K$ and \emph{chose} \texttt{cp.async} over TMA---an architectural
trade-off normally reserved for human kernel engineers. Listing~\ref{lst:veomni-dw1}
shows the resulting WGMMA mainloop.

\begin{lstlisting}[language=C++,basicstyle=\ttfamily\scriptsize,caption={\model-generated WGMMA mainloop for the VeOmni \texttt{dW1} weight-gradient kernel (excerpt). No library GEMM is invoked; the matrix multiply is a hand-written SM90 warp-group MMA.},label={lst:veomni-dw1}]
// SM90 warp-group MMA atom (MN-major operands, FP32 accumulate); NS=4 stages
using Atom = typename MmaSel<TA>::type;   // SM90_64x128x16_F32F16F16_SS
TiledMMA mma = make_tiled_mma(Atom{});
...
for (int kt = 0; kt < k_tiles; ++kt) {              // ragged-K mainloop
  cp_async_wait<NS - 2>(); __syncthreads();
  int kload = kt + (NS - 1);                        // prefetch NS-1 tiles ahead
  if (kload < k_tiles) {                            // predicated cp.async (4-stage)
    fillA(kload); copy_if(copyA, tApA, tAgA(_,_,_,kload), tAsA(_,_,_,smem_pipe_write));
    fillB(kload); copy_if(copyB, tBpB, tBgB(_,_,_,kload), tBsB(_,_,_,smem_pipe_write));
  }
  cp_async_fence(); smem_pipe_write = (smem_pipe_write + 1) % NS;
  warpgroup_arrive();
  cute::gemm(mma, tCrA(_,_,_,smem_pipe_read), tCrB(_,_,_,smem_pipe_read), tCrC);  // WGMMA
  warpgroup_commit_batch(); warpgroup_wait<0>();
  smem_pipe_read = (smem_pipe_read + 1) % NS;
}
\end{lstlisting}

\paragraph{SASS analysis in the trajectory.}
This run also exercises \model's \textbf{instruction-level} SASS tools
(Section~\ref{sec:sass-tools}) in a way kernel-level \texttt{ncu} cannot. Because the
LLM frequently emits code it \emph{believes} uses warp-group tensor cores, the
\textbf{WGMMAInstructionTool} disassembles each compiled \texttt{.so} and reports the
tensor-core instruction mix directly. Across the search it flagged \textbf{37
candidates} that compiled and ran but emitted zero tensor-core instructions
(\texttt{Found 0 TC ops (WGMMA:0 HMMA:0)})---scalar fallbacks that no utilization
metric would distinguish from a slow-but-real tensor-core kernel---and surfaced
generation regressions such as a candidate using the older Ampere \texttt{HMMA}
instructions with \textbf{250 register-spill instructions}
(\texttt{16 TC ops (HMMA:16), STL:125 LDL:125}). The decisive late iterations were
in turn driven by an occupancy diagnosis---\model's profiling reported the kernel
\emph{latency-bound at $18.8\%$ warp occupancy, limited by $168$ registers/thread and
shared-memory pressure}---which steered the search toward the smaller-footprint,
higher-occupancy $64{\times}128{\times}16$ configuration that crossed the Triton
baseline. The optimization is thus guided by instruction- and occupancy-level
evidence, not runtime counters alone. The complete \model-generated kernel---mainloop, prologue/epilogue, and host launch---is provided in Appendix~\ref{appendix:veomni-dw1-source}; Listing~\ref{lst:veomni-dw1} shows only the WGMMA mainloop excerpt.

\subsection{Energy-Aware Kernel Optimization: A Preliminary Study}
\label{sec:energy-extension}

Power is increasingly the limiting factor in AI scaling~\citep{iea2024electricity}, and
GPU energy is dominated by data movement rather than compute~\citep{horowitz2014computing}.
This makes energy a first-class measure of kernel efficiency---yet existing LLM-based kernel
agents target latency alone. \model is the
first CUDA kernel coding agent to optimize energy efficiency; we explore
whether its profiling-driven search can reduce kernel energy, and where that reduction is
separable from speedup.

\model's design extends naturally beyond speedup. We treat energy efficiency as a
\emph{secondary} objective under strict lexicographic priority: speedup always wins,
and energy is optimized only to break ties among speed-equivalent solutions, via a single
energy term added to the reward:
\begin{equation}
\label{eq:energy-reward}
r(n) = \ln(\text{speedup}(n)) + \varepsilon \cdot \ln(\text{energy\_reduction}(n)),
\quad \varepsilon \ll 1,
\end{equation}
for correct solutions ($\varepsilon$ small enough that any speedup gain dominates any
energy gain). The search, tools, and pipeline are otherwise unchanged.

\paragraph{Kernel-level energy model.} GPU energy is dominated by data movement, not
compute: a DRAM access costs $\sim$100--200$\times$ a register access~\citep{horowitz2014computing},
and the register file is the single largest dynamic-power component on modern
GPUs~\citep{kandiah2021accelwattch}. Device-level power sensors (\texttt{nvidia-smi})
are too coarse and noisy for per-kernel attribution on A100/H100~\citep{yang2024parttime},
so we instead estimate energy deterministically from \texttt{ncu} counters \model already
collects, as a weighted cost over four documented bottlenecks, with weights set by the
per-operation energy hierarchy of Horowitz~\citep{horowitz2014computing} and
AccelWattch~\citep{kandiah2021accelwattch}:
{\footnotesize
\begin{equation}
\label{eq:energy-proxy}
E_\text{proxy} = \underbrace{20\,\text{pJ/B}\cdot B_\text{DRAM}}_{\text{B1: DRAM bytes}}
  + \underbrace{100\,\text{pJ}\cdot S_\text{ld}\!\cdot\!\max(0,\tfrac{r}{4}\!-\!1)}_{\text{B2: uncoalesced}}
  + \underbrace{80\,\text{pJ}\cdot S_\text{local}}_{\text{B3: spill traffic}}
  + \underbrace{3\,\text{pJ}\cdot I}_{\text{B4: instructions}}
\end{equation}
}
where $B_\text{DRAM}$ is DRAM bytes read+written, $S_\text{ld}$ global-load sectors, $r$
sectors-per-request, $S_\text{local}$ local-memory (spill) sectors, and $I$ instructions
executed. The resulting \texttt{energy\_reduction} ratio (baseline/candidate cost) is
intrinsic to the kernel and independent of wall-clock time, so it rewards reductions in
hardware activity that timing alone cannot see. To confirm that proxy-driven choices
translate to real device energy, we separately measure winning kernels in millijoules
using a rigorous locked-clock, idle-subtracted NVML protocol. We also add four
\emph{energy-only} micro-profiling tools that flag latency-hidden energy waste (bank
conflicts, register spills, uncoalesced access, redundant barriers) the speed-only tools
ignore. Appendix~\ref{appendix:energy-methodology} details the NVML protocol, the energy
model, the four tools, and the literature mapping each to a documented bottleneck.

To isolate the energy-aware reward from speedup, we run
paired arms (i.e., speed-only versus energy-aware), and measure the winning kernels in millijoules under the rigorous protocol of Appendix~\ref{app:nvml}. Most energy savings coincide with speed savings---reducing data movement cuts both---so the speed-only search already captures them; the interesting case is where they decouple. On the \texttt{Swish} activation (KernelBench L1, pure
elementwise, memory-bound), both arms reach an identical $2.52\times$ speedup with
identical memory traffic (1 vectorized load + 1 store per thread), yet the energy-aware
kernel draws \textbf{1802\,mJ versus 2038\,mJ---11.6\% less dynamic energy}
(Table~\ref{tab:energy-extension}). The mechanism is visible in SASS: the energy-aware
kernel compiles to \textbf{56 instructions versus 216} for the same computation, having
chosen the fast reciprocal intrinsic (\texttt{\_\_fdividef}, one \texttt{MUFU}) over an
IEEE-accurate division (32 \texttt{FFMA} + branch-heavy refinement) and a read-only cache
load (\texttt{\_\_ldg}). Because the kernel is bandwidth-bound, those $\sim$160 extra
instructions hide under memory stalls and cost \emph{no} latency---but the SMs still
execute and draw power for them. The speed-only objective is blind to this; the energy
term is what selects the cheaper instruction stream. This is a concrete instance of energy
reducible at fixed speed; a systematic characterization is left to future work.

\begin{table}[t]
\centering
\small
\caption{Energy-aware matched-speed A/B on the \texttt{Swish} kernel (KernelBench L1).
Both arms reach identical speedup and memory traffic; dynamic energy is measured under the
locked-clock, idle-subtracted NVML protocol (Appendix~\ref{app:nvml}). SASS instruction
counts are for the vectorized kernel.}
\label{tab:energy-extension}
\begin{tabular}{@{}lcccc@{}}
\toprule
\textbf{Configuration} & \textbf{Speedup} & \textbf{Dynamic energy} & \textbf{SASS insns} & \textbf{$\Delta$ energy} \\
                       &                  & \textbf{(mJ, NVML)}     &                     &                 \\
\midrule
\model (speed-only)     & 2.52$\times$ & 2038 & 216 & --- \\
\model (energy-aware)   & 2.52$\times$ & \textbf{1802} & \textbf{56} & \textbf{$-11.6\%$} \\
\bottomrule
\end{tabular}
\end{table}

\section{Conclusion}

\model demonstrates that the key bottleneck in LLM-based GPU kernel optimization is not code generation capability but \emph{how profiling data reaches the LLM}. By encoding expert heuristics as pluggable micro-profiling tools---executable analysis functions that transform raw hardware counters into actionable natural language guidance---\model replaces the implicit reasoning required by prior systems with explicit, systematic diagnostic workflows. Controlled ablations establish a clear contribution hierarchy: micro-profiling tools provide the largest gain (125\% over raw metrics), with the counterintuitive finding that raw \texttt{ncu} metrics \emph{actively harm} optimization quality relative to no feedback at all---directly challenging systems that pass unstructured counters to LLMs; domain-adapted MCTS and proactive tool orchestration each contribute significant further gains. Together these yield geometric mean speedups of $2.42\times$/$4.69\times$/$5.30\times$ on KernelBench Levels 1/2/3 with 100\% task coverage, surpassing the prior SOTA on all levels. Beyond academic benchmarks, \model achieves 1.23$\times$ over an expert-optimized Triton baseline on a production MoE weight-gradient kernel (H100), generating a from-scratch raw-CUDA+CuTe Hopper WGMMA kernel, not a library call. \model's CUTLASS/CuTe integration---with LLM-autonomous code search and architecture constraint mining---further enables direct source-level tensor core code generation, grounding the LLM in authoritative library source rather than memorized patterns (Appendix~\ref{appendix:cutlass-case-studies} presents two complete trajectories). Finally, as the first CUDA kernel coding agent to optimize energy efficiency, \model reduces measured energy by $11.6\%$ at matched speed via lower-energy instruction selection---a preliminary result whose systematic characterization we leave to future work. \model has submitted 6 PRs to FlashInfer (1 merged, 5 under review), and scaling this into a sustained PR contribution pipeline is a key future direction.

\bibliographystyle{tmlr}
\bibliography{ifco}

\appendix

\section{Comparison with Concurrent Work}
\label{appendix:comparison}

Table~\ref{tab:landscape} provides an overview of concurrent LLM-based GPU kernel optimization systems: KernelBlaster~\citep{dong2026kernelblaster}, CudaForge~\citep{zhang2025cudaforge}, StitchCUDA~\citep{li2026stitchcuda}, KernelFoundry~\citep{wiedemann2026kernelfoundry}, and AVO~\citep{chen2026avo}. Cross-system comparisons warrant several caveats. First, although speedups are all measured relative to a PyTorch baseline on the same GPU, different architectures offer different optimization ceilings (e.g., Hopper tensor cores vs.\ Ampere), so absolute ratios are not directly comparable across hardware. Second, the systems differ in which KernelBench tasks they evaluate on; for instance, StitchCUDA trains on 80\% of KernelBench and evaluates on the remaining 20\%, so its numbers reflect in-distribution performance rather than zero-shot evaluation on the full benchmark. Third, the reported metric differs: \model and KernelBlaster report geometric mean speedup, whereas CudaForge reports arithmetic mean and StitchCUDA reports an end-to-end average. We plan to open-source \model's generated solutions for all 250 KernelBench tasks, pending approval.

\begin{table*}[t]
\centering
\footnotesize
\caption{Landscape of LLM-based GPU kernel optimization systems on KernelBench. Speedup is over PyTorch eager baseline on the same GPU. ``Tasks'' denotes the number of test tasks evaluated per level. Metric type: GeoMean (G), ArithMean (A).}
\label{tab:landscape}
\begin{tabular}{@{}llllccc>{\raggedright\arraybackslash}p{4.2cm}@{}}
\toprule
\textbf{System} & \textbf{LLM} & \textbf{GPU} & \textbf{Metric} & \textbf{L1} & \textbf{L2} & \textbf{L3} & \textbf{Notes} \\
\midrule
\textbf{\model (ours)} & Sonnet 4.6 & A100 & G & \textbf{2.42$\times$} & \textbf{4.69$\times$} & \textbf{5.30$\times$} & Full KernelBench (100/100/50 tasks) \\
KernelBlaster & GPT-4.1/5.0 & H100 & G & 1.43$\times$ & 2.50$\times$ & 1.50$\times$ & Full L1/L2; L3 subset (size unspecified) \\
CudaForge & o3 & RTX 6000 & A & 1.45$\times$ & 2.10$\times$ & 1.28$\times$ & Full KernelBench (100/100/50 tasks) \\
StitchCUDA & Qwen3-32B$^\ddagger$ & H200 & A & 3.54$\times$ & 1.82$\times$ & 1.50$\times$ & 20/20/10 test tasks; trained on remaining 80\% \\
KernelFoundry & GPT-4.1/o3 & RTX A6000 & -- & -- & 2.10$\times$ & -- & 40-task subset; MAP-Elites evolutionary search \\
AVO & Undisclosed & B200 & -- & -- & -- & -- & Attention kernels only (not KernelBench) \\
\bottomrule
\end{tabular}
\vspace{2pt}

{\footnotesize $^\ddagger$StitchCUDA uses GRPO-trained Qwen3-32B as Coder (trained on 80\% of KernelBench), GPT-5.2 as Planner/Verifier.}
\end{table*}

\section{Micro-Profiling Tool Specifications}
\label{appendix:tool-specs}

\begin{table*}[t]
\centering
\small
\caption{\model micro-profiling tool taxonomy and trigger conditions}
\label{tab:tools}
\begin{tabular}{@{}lllp{4.5cm}@{}}
\toprule
\textbf{Tool} & \textbf{Category} & \textbf{Bottleneck} & \textbf{Trigger Condition} \\
\midrule
TensorCoreUnderutilization & NCU & COMPUTE & TC util $<$ 10\% \\
TensorCoreModerateUtil & NCU & COMPUTE & TC util 10--50\% \\
WGMMAInstruction & SASS & COMPUTE & WGMMA/HMMA detected \\
RegisterSpillDetector & SASS & LATENCY & STL/LDL + bottleneck match \\
MemoryCoalescing & NCU & MEMORY & L1 hit $<$ 30\%, stalls $>$ 40\% \\
HighDRAMThroughput & NCU & MEMORY & DRAM throughput $>$ 80\% \\
OccupancyLimiter & NCU & OCCUPANCY & Occupancy $<$ 50\% \\
HighRegisterUsage & \texttt{ncu} & OCCUPANCY & Registers $>$ 96 \\
WarpStall & NCU & LATENCY & Dominant stall $>$ 40\% \\
BarrierStall & NCU & LATENCY & Barrier stall $>$ 30\% \\
LaunchOverhead & nsys & LAUNCH & Launch overhead $>$ 20\% \\
SyncOverhead & nsys & LAUNCH & Sync overhead $>$ 10\% \\
SmallKernel & \texttt{nsys} & LAUNCH & Kernel $<$ 10$\mu$s, launches $>$ 50 \\
MemoryTransfer & nsys & MEMORY & Transfer $>$ 15\% runtime \\
InterKernelGap & nsys & LAUNCH & Gap $>$ 1\% runtime \\
\bottomrule
\end{tabular}
\end{table*}

\subsection{Detailed Tool Formalizations}

\paragraph{WarpStallTool.}
Identifies \emph{why} warps fail to issue by finding the dominant stall reason among five \texttt{ncu} warp-stall counters, then emitting a fix tailored to that reason---the clearest instance of diagnostic reasoning over raw numbers. The \texttt{analyze} step dispatches on the dominant stall: memory-dependency $\rightarrow$ shared-memory caching and vectorized (\texttt{float4}) loads; barrier $\rightarrow$ warp-level primitives (\texttt{\_\_shfl\_sync}/\texttt{\_\_ballot\_sync}) in place of \texttt{\_\_syncthreads()}; long-scoreboard $\rightarrow$ async prefetch (\texttt{cp.async} on Ampere+, TMA on Hopper); branch-resolving $\rightarrow$ arithmetic predication.
\begin{align*}
    \mathcal{M}_{\text{req}} &= \{\texttt{mem\_dep}, \texttt{short\_scoreboard}, \texttt{long\_scoreboard},\\
    &\qquad \texttt{barrier}, \texttt{branch\_resolving}\} \quad (\text{stall \% per active warp}) \\
    \theta &= \{\text{high}{:}\,40\%,\ \text{critical}{:}\,60\%\} \\
    s^{*} &= \arg\max_{s \in \mathcal{M}_{\text{req}}} m.s \\
    \text{trigger}(m) &= \mathds{1}[\,m.s^{*} > \theta_{\text{high}}\,] \\
    \text{analyze} &\rightarrow (\text{severity by } m.s^{*};\ \text{rec}{:}\ \text{fix specific to } s^{*})
\end{align*}

\paragraph{WGMMAInstructionTool (SASS).}
Tensor core utilization is the single most impactful optimization for matrix-heavy GPU workloads. Critically, on Hopper GPUs, only \texttt{WGMMA} instructions can achieve peak throughput--older \texttt{HMMA} instructions cap at $\sim$63\% of peak, leaving 37\% performance on the table~\citep{spector2024thunderkittens}. This tool provides definitive tensor core detection by analyzing compiled SASS binary via \texttt{cuobjdump --dump-sass}, identifying not just \emph{whether} tensor cores are used but \emph{which generation} of instructions, enabling architecture-specific optimization guidance. Supports all tensor core instruction variants:

\begin{itemize}
    \item \textbf{WGMMA}: Hopper warpgroup MMA (sm\_90) -- FP16/BF16/FP8/INT8/TF32
    \item \textbf{HMMA}: Half-precision MMA (sm\_70+) -- FP16/BF16
    \item \textbf{IMMA/DMMA/BMMA}: Specialized MMA -- INT8/INT4, FP64, INT1
\end{itemize}

Additionally detects cuBLAS usage via \texttt{nm -D} symbol lookup (cuBLAS auto-selects tensor core kernels on Volta+ GPUs). The two-stage approach: (1) fast \texttt{nm -D} check for library calls, (2) full SASS disassembly for user-written kernels.

\begin{align*}
    \mathcal{M}_{\text{req}} &= \{\texttt{sass\_analysis}\} \\
    \text{tc\_count} &= \texttt{wgmma} + \texttt{hmma} + \texttt{imma} + \texttt{dmma} + \texttt{bmma} \\
    \text{trigger}(m) &= \mathds{1}[\text{tc\_count} > 0 \lor \texttt{uses\_cublas}] \\
    \text{analyze} &\rightarrow (\text{which TC generation is emitted; recommend the}\\
    &\qquad \text{architecture-appropriate path, e.g.\ WGMMA on Hopper})
\end{align*}

\paragraph{RegisterSpillDetector (SASS).}
Detection is multi-signal: from the SASS binary it counts \texttt{STL} (store local) and \texttt{LDL} (load local) instructions, then flags a spill ($\texttt{stl\_count}+\texttt{ldl\_count} > 0$) as harmful only when one of three conditions holds, with the exact cutoffs below:

\begin{itemize}
    \item \textbf{Case A (Latency-bound)}: long-scoreboard stall $>15\%$ \emph{and} compute throughput $<50\%$ \emph{and} memory throughput $<50\%$ (spills add latency while neither roofline is saturated).
    \item \textbf{Case B (Memory-bound by spills)}: memory throughput $>60\%$ \emph{and} local-memory traffic ratio $>30\%$ (spills consume bandwidth needed for useful data).
    \item \textbf{Case C (Occupancy-limited)}: registers are the binding occupancy limit \emph{and} the scheduler is starved (issue activity $<30\%$ \emph{or} eligible warps/cycle $<0.6$).
\end{itemize}

\begin{align*}
    \mathcal{M}_{\text{req}} &= \{\texttt{stl\_count}, \texttt{ldl\_count}, \texttt{long\_scoreboard}, \texttt{compute\_tput},\\
    &\qquad \texttt{memory\_tput}, \texttt{local\_traffic\_ratio}, \texttt{occupancy\_limit\_regs},\\
    &\qquad \texttt{issue\_active}, \texttt{eligible\_warps}\} \\
    \text{trigger}(m) &= \mathds{1}[\texttt{stl\_count}+\texttt{ldl\_count} > 0] \;\land\; (\text{A} \lor \text{B} \lor \text{C}) \\
    \text{analyze} &\rightarrow (\text{severity by case; rec}{:}\ \text{reduce register pressure}\\
    &\qquad \text{via \texttt{\_\_launch\_bounds\_\_}, \texttt{-maxrregcount}, or kernel split})
\end{align*}

\paragraph{TensorCoreUnderutilizationTool.}
Flags compute-bound kernels that run on CUDA cores instead of tensor cores. As a 5-tuple:
\begin{align*}
    \mathcal{M}_{\text{req}} &= \{\texttt{tensor\_core\_util}\} \\
    \theta &= \{\text{low}{:}\,10\%,\ \text{moderate}{:}\,50\%\} \\
    \text{trigger}(m) &= \mathds{1}[m.\texttt{tensor\_core\_util} < \theta_{\text{low}}] \\
    \text{analyze} &\rightarrow (\text{severity}{:}\ \textsc{critical},\\
    &\qquad \text{rec}{:}\ \text{``rewrite the matmul as a cuBLAS/CUTLASS GEMM''})
\end{align*}

\paragraph{LaunchOverheadDetector.}
Flags kernels whose end-to-end time is dominated by launch overhead rather than compute. As a 5-tuple:
\begin{align*}
    \mathcal{M}_{\text{req}} &= \{\texttt{launch\_overhead\_pct}\ (p_L),\ \texttt{avg\_launch\_ns},\ \texttt{kernel\_launches}\} \\
    \theta &= \{\text{high}{:}\,20\%,\ \text{critical}{:}\,40\%\} \\
    \text{trigger}(m) &= \mathds{1}[\,p_L > \theta_{\text{high}}\,] \\
    \text{analyze} &\rightarrow (\text{severity by } p_L,\ \text{rec}{:}\ \text{``CUDA Graph capture or kernel fusion''},\\
    &\qquad \text{est.\ speedup} = (1 - p_L/100)^{-1})
\end{align*}

The five tools above are representative; we refer readers to the open-source release for the complete set and their full definitions.

\subsection{Defining a New Tool}
\label{app:tool-registration}

\model's tools are pluggable: a new heuristic is added by subclassing
\texttt{ProfilingTool}, declaring its required \texttt{ncu} metrics and thresholds,
implementing \texttt{analyze()}, and registering it with the
\texttt{@ToolRegistry.register} decorator---after which the proactive orchestrator
invokes it automatically (no changes to the core system). Listing~\ref{lst:tool-reg}
shows a complete tool (the register-pressure detector); \texttt{analyze()} returns a
structured \texttt{ToolOutput} carrying the severity, root cause, and ranked
recommendations that become the natural-language feedback injected into the prompt.

\begin{lstlisting}[language=Python,caption={Defining and registering a micro-profiling tool. The \texttt{analyze} method returns a structured \texttt{ToolOutput} (severity, root cause, ranked recommendations) that the orchestrator renders into prompt feedback.},label={lst:tool-reg}]
@ToolRegistry.register
class HighRegisterUsageTool(ProfilingTool):
    name = "high_register_usage"
    priority = 70                 # higher runs first
    REG_WARNING, REG_CRITICAL = 96, 128

    @property
    def required_metrics(self) -> List[str]:
        return ["launch__registers_per_thread"]

    def analyze(self, kernel_name: str,
                metrics: Dict[str, Any]) -> ToolOutput:
        regs = self.get_metric(metrics,
                 "launch__registers_per_thread", default=32)
        if regs < self.REG_WARNING:
            return ToolOutput(triggered=False)
        critical = regs >= self.REG_CRITICAL
        return ToolOutput(
            triggered=True,
            severity=Severity.HIGH if critical else Severity.MEDIUM,
            title="High Register Usage",
            summary=f"{regs:.0f} registers/thread - may spill to local memory",
            root_cause="Register demand exceeds the physical file; spills to "
                       "local memory are 10-100x slower than registers.",
            recommendations=[
                "Use __launch_bounds__ to guide the compiler",
                "Cap registers with -maxrregcount",
                "Split the kernel or move data to shared memory"],
            expected_improvement="2-10x if eliminating spilling")
\end{lstlisting}

Tools span a wide complexity range under the same \texttt{analyze}$\rightarrow$\texttt{ToolOutput} interface. Listing~\ref{lst:tool-reg} is a minimal single-threshold detector; Listing~\ref{lst:warpstall} shows production-grade diagnostic logic---\texttt{WarpStallTool} reads five warp-stall counters, identifies the \emph{dominant} stall reason via \texttt{argmax}, and dispatches a root cause and recommendations tailored to that reason (the same expert ``which stall, and why'' reasoning a human applies to \texttt{ncu} output). We reproduce its complete implementation.

\begin{lstlisting}[language=Python,basicstyle=\ttfamily\scriptsize,caption={\texttt{WarpStallTool}: complete implementation. It selects the dominant stall reason among five \texttt{ncu} counters via \texttt{argmax} and returns a reason-specific root cause and recommendations.},label={lst:warpstall}]
@ToolRegistry.register
class WarpStallTool(ProfilingTool):
    name = "warp_stall_analyzer"
    description = "Analyzes warp stall reasons to identify bottlenecks"
    version = "1.0.0"
    priority = 80

    # Thresholds
    STALL_HIGH = 40.0      # >40% stalls on one reason is significant
    STALL_CRITICAL = 60.0  # >60% is critical

    @property
    def required_metrics(self) -> List[str]:
        return [
            "smsp__warp_issue_stalled_memory_dependency_per_warp_active.pct",
            "smsp__warp_issue_stalled_short_scoreboard_per_warp_active.pct",
            "smsp__warp_issue_stalled_long_scoreboard_per_warp_active.pct",
            "smsp__warp_issue_stalled_barrier_per_warp_active.pct",
            "smsp__warp_issue_stalled_branch_resolving_per_warp_active.pct"
        ]

    def analyze(self, kernel_name: str, metrics: Dict[str, Any]) -> ToolOutput:
        # Get all stall metrics
        stalls = {
            "memory_dependency": self.get_metric(metrics,
                "smsp__warp_issue_stalled_memory_dependency_per_warp_active.pct", default=0),
            "short_scoreboard": self.get_metric(metrics,
                "smsp__warp_issue_stalled_short_scoreboard_per_warp_active.pct", default=0),
            "long_scoreboard": self.get_metric(metrics,
                "smsp__warp_issue_stalled_long_scoreboard_per_warp_active.pct", default=0),
            "barrier": self.get_metric(metrics,
                "smsp__warp_issue_stalled_barrier_per_warp_active.pct", default=0),
            "branch_resolving": self.get_metric(metrics,
                "smsp__warp_issue_stalled_branch_resolving_per_warp_active.pct", default=0)
        }

        # Find the dominant stall reason
        dominant_stall = max(stalls, key=stalls.get)
        dominant_value = stalls[dominant_stall]

        # Check if any stall is significant
        if dominant_value < self.STALL_HIGH:
            return ToolOutput(triggered=False)

        is_critical = dominant_value >= self.STALL_CRITICAL
        severity = Severity.HIGH if is_critical else Severity.MEDIUM

        # Get stall-specific analysis
        root_cause = self._get_stall_root_cause(dominant_stall)
        recommendations = self._get_stall_recommendations(dominant_stall, stalls)

        return ToolOutput(
            triggered=True,
            severity=severity,
            title=f"High Warp Stalls ({dominant_stall.replace('_', ' ').title()})",
            summary=f"{dominant_value:.1f}% of warps stalled on {dominant_stall}",
            metrics_observed={
                "Memory dependency stalls": f"{stalls['memory_dependency']:.1f}%",
                "Short scoreboard stalls": f"{stalls['short_scoreboard']:.1f}%",
                "Long scoreboard stalls": f"{stalls['long_scoreboard']:.1f}%",
                "Barrier stalls": f"{stalls['barrier']:.1f}%",
                "Branch resolving stalls": f"{stalls['branch_resolving']:.1f}%",
                "Dominant stall": dominant_stall
            },
            metrics_thresholds={"Any stall type": f"<{self.STALL_HIGH}% (target)"},
            root_cause=root_cause,
            recommendations=recommendations,
            expected_improvement="1.5-3x depending on stall type"
        )

    def _get_stall_root_cause(self, stall_type: str) -> str:
        causes = {
            "memory_dependency": "Warps wait for memory operations to complete "
                "(memory-bound): 400+ cycle global latency, uncoalesced accesses, "
                "or insufficient parallelism to hide latency.",
            "short_scoreboard": "Warps wait for shared-memory/texture (MIO) ops: "
                "shared-memory bank conflicts, texture cache misses, sync overhead.",
            "long_scoreboard": "Warps wait for long-latency ops tracked by the long "
                "scoreboard: outstanding global loads, texture fetches, atomics.",
            "barrier": "Warps wait at __syncthreads() for the block to catch up: "
                "load imbalance, too many sync points, or divergent execution times.",
            "branch_resolving": "Warps wait for branch conditions to resolve: "
                "data-dependent or thread-ID-dependent divergent paths."
        }
        return causes.get(stall_type, "Unknown stall type")

    def _get_stall_recommendations(self, stall_type: str, all_stalls: Dict) -> List[str]:
        base_recs = {
            "memory_dependency": [
                "Use shared memory to cache frequently accessed data",
                "Implement vectorized loads (float4) for better bandwidth",
                "Ensure coalesced memory access patterns",
                "Increase occupancy to hide memory latency (more warps)",
                "Use __ldg() for read-only data (uses texture cache)",
                "Consider software pipelining to overlap compute and memory"
            ],
            "short_scoreboard": [
                "Check for shared memory bank conflicts",
                "Reduce shared memory accesses if possible",
                "Use padding to avoid bank conflicts: `__shared__ float A[32][33]`",
                "Consider using registers instead of shared memory",
                "Profile with ncu --set full to see bank conflict counts"
            ],
            "long_scoreboard": [
                "Similar to memory dependency - optimize memory access",
                "Use async copy for prefetching if on Ampere+ (cp.async)",
                "Consider TMA on Hopper for bulk transfers",
                "Increase warp-level parallelism"
            ],
            "barrier": [
                "Reduce number of __syncthreads() calls",
                "Use warp-level primitives (__shfl, __ballot) when possible",
                "Balance workload across threads in a block",
                "Consider warp-synchronous programming for small data",
                "NEVER put __syncthreads() inside conditional blocks"
            ],
            "branch_resolving": [
                "Reorganize data to minimize branch divergence",
                "Use arithmetic instead of branches: result = cond*v1 + (1-cond)*v2",
                "Sort data to group similar computation paths",
                "Use predication for short conditional blocks",
                "Consider separate kernels for different code paths"
            ]
        }
        return base_recs.get(stall_type, ["Profile further to identify root cause"])
\end{lstlisting}

\section{Micro-Profiling Tool Effectiveness: KernelBench and Production Kernels}
\label{appendix:kernelbench-ablation}

We isolate the contribution of \model's micro-profiling tools with two controlled
with-tools-versus-without-tools ablations: one on 42 KernelBench tasks
(\S\ref{app:ablation-kernelbench}) and one on six production Triton inference kernels
(\S\ref{app:ablation-production}), confirming the effect holds on both benchmark and
real serving workloads.

\subsection{KernelBench Tasks}
\label{app:ablation-kernelbench}

We conduct a controlled ablation on the 42-task subset (Section~\ref{sec:component-ablation}) to isolate the contribution of micro-profiling tools. Three configurations compare progressive levels of profiling feedback:

\begin{itemize}
    \item \textbf{No Feedback}: The LLM receives only source code and compilation/correctness errors (\texttt{--skip-profiling}). Represents naive LLM retry.
    \item \textbf{Raw \texttt{ncu} Metrics}: Full NCU counter dump ($\sim$50 raw metric values) appended to the prompt with no interpretation or bottleneck classification (\texttt{--raw-metrics-only}).
    \item \textbf{Ours}: Raw \texttt{ncu} metrics plus all \model micro-profiling tools (Stage~1 roofline classification + Stage~2 filtered tool analysis with structured optimization guidance).
\end{itemize}

All conditions use Sonnet 4.6 on A100 GPUs with 15 seeds per task (5 temperatures $\times$ 3 independent rounds), 30 iterations, 2 candidates per iteration, and greedy search. We report the best speedup across all seeds for each task-configuration pair, with outlier speedups capped to limit disproportionate influence.

\paragraph{Task IDs.} The 42 KernelBench tasks used in this ablation are:
\begin{itemize}[nosep,leftmargin=*]
    \item \textbf{Level 1} (19): 1, 6, 7, 8, 10, 11, 12, 13, 14, 15, 16, 17, 18, 23, 24, 26, 40, 88, 97
    \item \textbf{Level 2} (16): 3, 9, 12, 14, 18, 19, 22, 29, 30, 33, 34, 36, 37, 39, 40, 41
    \item \textbf{Level 3} (7): 28, 30, 31, 32, 43, 44, 50
\end{itemize}

Table~\ref{tab:ablation-correctness} shows the fraction of tasks where each configuration produces a correct kernel faster than the PyTorch eager baseline.

\begin{table}[ht]
\centering
\small
\caption{Correctness ablation: fraction of tasks solved (speedup $> 1\times$).}
\label{tab:ablation-correctness}
\begin{tabular}{@{}lcc@{}}
\toprule
\textbf{Configuration} & \textbf{Tasks Solved} & \textbf{Pass Rate (fast\_1)} \\
\midrule
No Feedback & 32 / 42 & 76.2\% \\
Raw \texttt{ncu} Metrics & 36 / 42 & 85.7\% \\
\textbf{Ours} & \textbf{42 / 42} & \textbf{100.0\%} \\
\bottomrule
\end{tabular}
\end{table}

Ours achieves a 100\% solve rate across all 42 tasks, solving 31\% more tasks than No Feedback and 17\% more than Raw \texttt{ncu} Metrics. On several tasks, only Ours achieves substantial speedup (e.g., L2 task~40: 7.5$\times$ vs 1.0$\times$/1.0$\times$; L2 task~41: 5.8$\times$ vs 1.0$\times$/1.0$\times$), demonstrating that profiling-guided insight unlocks optimizations inaccessible to unguided approaches. The L2 task~41 trajectory (Appendix~\ref{appendix:case-l2-41}) provides a detailed walkthrough of how tool guidance produces its $12.1\times$ final speedup---the $2.8\times$ single-step improvement, where the agent acts on the tools' top-ranked directives (epilogue fusion plus BF16 tensor cores), exemplifies the mechanism by which structured feedback unlocks optimization paths invisible to unguided search.

Table~\ref{tab:ablation-speedup} compares optimization quality across configurations.

\begin{table}[ht]
\centering
\small
\caption{Speedup ablation (outliers capped, unsolved tasks scored as 1.0$\times$).}
\label{tab:ablation-speedup}
\begin{tabular}{@{}lccc@{}}
\toprule
\textbf{Configuration} & \textbf{Mean} & \textbf{Geo Mean} & \textbf{Win Rate} \\
\midrule
No Feedback & 6.59$\times$ & 3.35$\times$ & 10/42 (24\%) \\
Raw \texttt{ncu} Metrics & 3.45$\times$ & 1.77$\times$ & 3/42 (7\%) \\
\textbf{Ours} & \textbf{6.91$\times$} & \textbf{4.00$\times$} & \textbf{25/42 (60\%)} \\
\bottomrule
\end{tabular}
\end{table}

Table~\ref{tab:ablation-fastp} shows the fraction of tasks exceeding progressively higher speedup thresholds. Ours dominates across all thresholds, with the gap widening at higher values.

\begin{table}[ht]
\centering
\small
\caption{fast\_p: fraction of tasks with speedup $> p$.}
\label{tab:ablation-fastp}
\begin{tabular}{@{}lccc@{}}
\toprule
\textbf{Threshold $p$} & \textbf{No Feedback} & \textbf{Raw \texttt{ncu}} & \textbf{Ours} \\
\midrule
$>1\times$ & 0.762 & 0.857 & \textbf{1.000} \\
$>1.5\times$ & 0.548 & 0.286 & \textbf{0.667} \\
$>2\times$ & 0.524 & 0.238 & \textbf{0.619} \\
$>3\times$ & 0.524 & 0.238 & \textbf{0.595} \\
$>5\times$ & 0.405 & 0.167 & \textbf{0.524} \\
$>10\times$ & 0.167 & 0.095 & \textbf{0.214} \\
\bottomrule
\end{tabular}
\end{table}

We use the Wilcoxon signed-rank test~\citep{hollander2014nonparametric}, a non-parametric paired test that makes no distributional assumptions, to evaluate whether configurations differ in optimization quality. For each task, we take the best speedup (with outliers capped) as that configuration's score; unsolved tasks are scored as 1.0$\times$. The test is one-sided ($H_1$: first configuration produces higher speedups); $n$ reports the number of non-zero differences (Wilcoxon excludes tied pairs). Results (Table~\ref{tab:ablation-wilcoxon-main}) confirm the ordering Ours $>$ No Feedback $>$ Raw \texttt{ncu}.

We highlight three key observations from the ablation results.

\paragraph{Full pipeline is necessary for reliable optimization.} Ours solves 100\% of tasks--the structured profiling pipeline consistently identifies actionable bottlenecks that the model can address.

\paragraph{Raw metrics without interpretation are harmful.} Raw \texttt{ncu} Metrics performs significantly \emph{worse} than No Feedback ($p = 0.0007$). This condition passes ${\sim}50$ raw NCU counters without interpretation--comparable to CudaForge~\citep{zhang2025cudaforge}, which presents statistically pre-filtered metrics as verbatim name-value pairs. Unstructured metric dumps appear to distract the model: it attempts to optimize counters that do not translate to wall-clock improvements, or misinterprets raw values without roofline context. This validates \model's core design principle that \emph{interpretation}, not mere data access, drives optimization quality.

\paragraph{The interpretation layer is the key differentiator.} The gap from Raw \texttt{ncu} to Ours (raw $\to$ analyzed) is larger than from No Feedback to Ours (none $\to$ analyzed). The value lies not in \emph{collecting} profiling data, but in \emph{presenting it as actionable, structured optimization guidance} via the Stage~1/Stage~2 pipeline. This reveals that \model's tool design is not simply ``give the LLM more data''--it is specifically about \emph{structured interpretation} of profiling data. Raw metrics actively degrade performance because the LLM lacks the domain context to translate hardware counters into optimization actions; only the micro-profiling tools' detect-analyze-recommend pattern bridges this gap reliably.

\paragraph{Profiling prevents correct-but-slow plateaus.} Trajectory analysis reveals that \model never produces a correct kernel that fails to exceed baseline (0/62 trajectories stuck at 1.0$\times$), while No Feedback gets stuck in 10\% of cases. When both configurations recover from an initial compilation error, No Feedback often produces a kernel that merely matches baseline performance and cannot improve further without knowing what to optimize. \model immediately receives structured guidance (e.g., ``reduce register usage from 168 to $<$64 per thread'') and follows a clear optimization path. This explains why \model achieves higher fast\_1 despite similar raw correctness rates: profiling does not fix bugs, but it prevents the model from settling on correct-but-slow solutions by always providing an actionable optimization target.

\subsection{Per-Tool Impact Analysis}
\label{appendix:per-tool-impact}

Per-tool coverage and hit rates are reported in Table~\ref{tab:tool-impact-main} (main body). Below we provide additional per-bottleneck breakdowns.

Table~\ref{tab:bottleneck-optimization} shows the dominant optimization action triggered by each bottleneck type, measured as the single largest improvement per task.

\begin{table}[ht]
\centering
\small
\caption{Most impactful optimization action per bottleneck type across 42 tasks.}
\label{tab:bottleneck-optimization}
\begin{tabular}{@{}llccc@{}}
\toprule
\textbf{Bottleneck} & \textbf{Primary Action} & \textbf{Tasks} & \textbf{Avg.} & \textbf{Max} \\
\midrule
Compute & Tensor Core switch & 9 & 6.25$\times$ & 9.6$\times$ \\
Latency & Register optimization & 4 & 6.14$\times$ & 10.6$\times$ \\
Memory & Coalescing/Vectorize & 6 & 3.48$\times$ & 8.6$\times$ \\
\bottomrule
\end{tabular}
\end{table}

Three observations emerge:

\paragraph{Tensor Core switch is the highest-impact single action.} For compute-bound kernels, the tools identify that the LLM-generated CUDA kernel uses CUDA cores instead of Tensor Cores and recommend switching to cuBLAS/CUTLASS. This single recommendation produces an average 6.25$\times$ gain across 9 tasks (max 9.6$\times$). Without the tool's quantitative signal (e.g., ``TC utilization: 0.7\%, threshold $>$10\%''), the LLM has no way to know its manual kernel is orders of magnitude slower than the hardware's peak capability. The Task~41 trajectory (Appendix~\ref{appendix:case-l2-41}) traces this pattern end-to-end: at iteration~13 the agent acts on the tools' top directives---fusing the epilogue and moving to a BF16 tensor-core GEMM---for a $4.11\times \to 11.49\times$ leap; the full generated solution is reproduced in the same appendix.

\paragraph{Register spill detection has the highest per-fire reliability.} RegisterSpill fires in only 16/42 tasks but converts to $>$1.5$\times$ improvement 18.2\% of the time---the highest among all tools. The diagnostic ``register spill detected: 168 registers/thread, spilling to local memory'' is highly actionable: the LLM can directly reduce register pressure by restructuring loops or reducing live variables. Example: L1\_1 goes from 1.0$\times$ to 10.6$\times$ after the LLM eliminates register spills.

\paragraph{Memory optimizations provide broad but incremental gains.} MemoryCoalescing and Vectorization fire in 37--41 of 42 tasks but have low hit rates (4--6\%). Memory access patterns are the most common bottleneck on A100 but are harder for the LLM to fix correctly---coalescing requires non-trivial data layout changes, while vectorization requires careful alignment handling.

\subsection{Production Inference Kernels}
\label{app:ablation-production}

To confirm the effect holds beyond KernelBench, we repeat the with-tools-versus-without-tools ablation on six representative Triton inference kernels covering common transformer serving patterns (NVIDIA A100). The kernels are: \texttt{layer\_norm} (fused forward layer normalization with affine transform), \texttt{vec\_matmul} (single-vector matrix multiplication for decoding), \texttt{persistent\_vec\_matmul} (persistent-loop variant that reuses thread blocks across output tiles), \texttt{gemm\_split\_k} (split-K GEMM with atomic accumulation for high-parallelism reduction), \texttt{fused\_moe} (fused Mixture-of-Experts token routing and batched expert GEMM from vLLM), and \texttt{kernel\_paged\_attention} (paged multi-head attention with online softmax over non-contiguous KV-cache blocks from vLLM, supporting GQA). Both configurations receive identical \texttt{ncu}/\texttt{nsys}/SASS raw metrics; the only difference is whether the tool-generated guidance is included in the LLM prompt.

\begin{table}[ht]
\centering
\small
\caption{Triton inference kernel ablation: tools enable better speedup in fewer iterations.}
\label{tab:ironbench-ablation}
\begin{tabular}{@{}lcccc@{}}
\toprule
& \multicolumn{2}{c}{\textbf{With Tools}} & \multicolumn{2}{c}{\textbf{Without Tools}} \\
\cmidrule(lr){2-3} \cmidrule(lr){4-5}
\textbf{Kernel} & \textbf{Speedup} & \textbf{Iters} & \textbf{Speedup} & \textbf{Iters} \\
\midrule
\texttt{layer\_norm}              & 1.70$\times$ & 6  & 1.24$\times$ & 10 \\
\texttt{vec\_matmul}              & 2.03$\times$ & 4  & 1.53$\times$ & 10 \\
\texttt{persistent\_vec\_matmul}  & 1.44$\times$ & 6  & 1.49$\times$ & 4  \\
\texttt{gemm\_split\_k}           & 2.13$\times$ & 5  & 1.39$\times$ & 2  \\
\texttt{fused\_moe}               & 3.26$\times$ & 2  & Failed       & 10 \\
\texttt{kernel\_paged\_attention} & 1.98$\times$ & 7  & 1.86$\times$ & 10 \\
\midrule
\textbf{Geo Mean}                 & \textbf{2.02$\times$} & \textbf{5.0} & 1.49$\times$ & 7.2 \\
\bottomrule
\end{tabular}
\end{table}

Tools achieved 36\% higher geometric mean speedup (2.02$\times$ vs 1.49$\times$) in 31\% fewer iterations (5.0 vs 7.2). \texttt{fused\_moe} failed entirely without tools after 10 iterations but achieved 3.26$\times$ with tools in just 2 iterations. The \texttt{vec\_matmul} case illustrates the ``fast but wrong'' trap: without tools, the LLM repeatedly generated incorrect solutions achieving 1.69--1.80$\times$ speedups that failed correctness, unable to diagnose why faster attempts broke. With tool guidance identifying the bottleneck (``shared memory limiting occupancy to 2\%''), it switched to warp-level reduction (\texttt{\_\_shfl\_down\_sync}) and found a correct 2.03$\times$ solution in 4 iterations.

\section{Search Memory Ablation}
\label{appendix:memory-ablation}

We conduct a controlled ablation to isolate the contribution of search memory--\model's cross-iteration learning mechanism described in Section~\ref{sec:session-memory}. Two configurations compare the full \model pipeline with and without memory:

\begin{itemize}
    \item \textbf{Memory ON (Treatment)}: Full \model pipeline with session memory extraction. Accumulated findings (\texttt{errors\_and\_corrections}, \texttt{successful\_patterns}, \texttt{search\_findings}) are injected into subsequent iterations.
    \item \textbf{Memory OFF (Control)}: Same pipeline with \texttt{--disable-session-memory}. Each iteration sees only the problem spec and profiling feedback from the parent node. Represents memoryless search.
\end{itemize}

All conditions use Sonnet 4.6 on H100 GPUs with 25 iterations, 2 candidates per iteration, greedy search, and temperature 0.2. Both conditions run full \texttt{ncu} profiling and the profiling agent--the sole variable is whether cross-iteration findings are extracted and injected. Outlier speedups are capped at 25$\times$ (consistent with Appendix~\ref{appendix:kernelbench-ablation}). This ablation uses the same 42-task subset as the other component ablations (Section~\ref{sec:component-ablation}).

\subsection{Search Memory Does Not Change Final Speedup}

Table~\ref{tab:memory-ablation-speedup} reports the best speedup achieved by each condition (capped at 25$\times$).

\begin{table}[ht]
\centering
\small
\caption{Search memory ablation: optimization quality.}
\label{tab:memory-ablation-speedup}
\begin{tabular}{@{}lcc@{}}
\toprule
\textbf{Metric} & \textbf{Memory ON} & \textbf{Memory OFF} \\
\midrule
Geometric mean speedup & 1.818$\times$ & 1.716$\times$ \\
Arithmetic mean speedup & 2.436$\times$ & 2.493$\times$ \\
Win / Loss / Tie & 9 / 7 / 20 & --- \\
\midrule
Wilcoxon $p$ (one-sided) & \multicolumn{2}{c}{0.181 (n.s.)} \\
Wilcoxon on $\log$ speedup & \multicolumn{2}{c}{0.181 (n.s.)} \\
\bottomrule
\end{tabular}
\end{table}

The Wilcoxon signed-rank test finds no significant difference in final optimization quality ($z = 0.911$, $p = 0.181$, $n = 36$). The geometric mean ratio (ON/OFF) is $1.06\times$--a small positive effect that does not reach significance due to high per-task variance. The 90\% CI for the ratio ($[0.936, 1.200]$) bounds any quality difference to at most 20\%, so the flat ceiling is a bounded result rather than merely underpowered.

\subsection{Search Memory Speeds Early Convergence}

\begin{wraptable}{r}{0.47\textwidth}
\centering
\small
\setlength{\tabcolsep}{4.5pt}
\caption{Convergence: geometric mean best@$N$ (capped at 25$\times$). Wilcoxon on $\log(\text{best@}N)$.}
\label{tab:memory-ablation-convergence}
\begin{tabular}{@{}lccccc@{}}
\toprule
\textbf{$N$} & \textbf{ON} & \textbf{OFF} & \textbf{Ratio} & \textbf{$z$} & \textbf{$p$} \\
\midrule
5  & 1.011$\times$ & 0.982$\times$ & 1.030 & 1.441 & 0.075$^\dagger$ \\
10 & 1.418$\times$ & 1.358$\times$ & 1.044 & 1.461 & 0.072$^\dagger$ \\
15 & 1.774$\times$ & 1.604$\times$ & 1.106 & 1.037 & 0.150 \\
25 & 1.782$\times$ & 1.656$\times$ & 1.076 & 0.833 & 0.203 \\
50 & 1.875$\times$ & 1.723$\times$ & 1.088 & 0.952 & 0.170 \\
\bottomrule
\end{tabular}

{\scriptsize $^\dagger$Marginal ($p < 0.10$). best@$N$.}
\end{wraptable}

While final speedup is equivalent, memory shows a marginal advantage in \emph{early} convergence. Table~\ref{tab:memory-ablation-convergence} reports best-so-far speedup at progressive search depths.

Memory shows its strongest effect in the first 10 iterations ($p = 0.072$), with treatment winning on 15/36 tasks versus 10/36 for control at $N = 10$ (mean $\log(T/C) = +0.043$). This advantage diminishes at deeper search depths, consistent with memory's role in avoiding redundant early exploration--memory prevents re-exploration of dead-end strategies during early iterations, but sufficiently long search compensates for the lack of cross-iteration learning. Its primary contribution is thus reaching good solutions faster rather than finding better ones given unlimited time. In practice this is the useful regime: under a fixed iteration budget--where users want the best kernel within a time constraint rather than the absolute best given unlimited search--the faster early convergence translates to reduced wall-clock time at no quality cost.

\section{Search Strategy Ablation: MCTS vs Greedy}
\label{appendix:search-strategy-ablation}

We conduct a controlled ablation to isolate the contribution of MCTS tree search versus greedy search. \model is the only LLM-based CUDA optimization agent to employ structured tree search; all related systems use flat iterative refinement or evolutionary population-based search. This ablation tests whether principled exploration via UCT selection produces better optimizations than pure greedy exploitation under a fixed compute budget of 60 candidates (30 iterations $\times$ 2 candidates per expansion).

Both conditions use the full \model profiling pipeline (Stage~1 roofline + Stage~2 filtered tools), session memory, Sonnet 4.6 on A100 GPUs, and identical hyperparameters: 15 seeds per task (5 temperatures $\times$ 3 rounds), 30 iterations, 2 candidates per iteration. The sole variable is the search strategy:

\begin{itemize}[nosep]
    \item \textbf{Greedy}: Each iteration expands only the current best node, with fresh restart after consecutive failures. Repair targets are selected via a composite priority \emph{(execution stage, error magnitude, speedup)}--kernels that execute but produce wrong output rank above compilation failures, with smaller numerical errors and higher partial speedups preferred as tiebreakers, concentrating repair budget on candidates closest to correctness.
    \item \textbf{MCTS}: Tree search with progressive widening and UCT selection ($c = \sqrt{2}$). Terminal marking prunes dead-end branches.
\end{itemize}

Table~\ref{tab:search-strategy-summary} reports aggregate metrics on the 42 KernelBench tasks. MCTS achieves a 26\% higher geometric mean speedup (4.60$\times$ vs 3.65$\times$) and wins on 29 tasks while greedy wins on only 8. The Wilcoxon signed-rank test rejects the null hypothesis at $p = 0.004$, providing strong evidence that MCTS produces higher-quality optimizations under matched compute budgets.

\begin{table}[ht]
\centering
\small
\caption{Search strategy ablation: MCTS vs Greedy (42 KernelBench tasks).}
\label{tab:search-strategy-summary}
\begin{tabular}{@{}lcc@{}}
\toprule
\textbf{Metric} & \textbf{Greedy} & \textbf{MCTS} \\
\midrule
Geometric mean speedup & 3.65$\times$ & \textbf{4.60$\times$} \\
Median speedup & 4.17$\times$ & \textbf{6.09$\times$} \\
fast\_p ($>2\times$) & 59.5\% & \textbf{70.3\%} \\
fast\_p ($>5\times$) & 48.6\% & \textbf{59.5\%} \\
Head-to-head wins & 8 & \textbf{29} \\
Wilcoxon $p$ (one-sided) & \multicolumn{2}{c}{0.004**} \\
\bottomrule
\end{tabular}
\end{table}

Table~\ref{tab:search-strategy-fastp} shows the fraction of tasks exceeding progressively higher speedup thresholds. MCTS dominates across all thresholds, with the gap largest at moderate speedups (1.5--5$\times$) where MCTS solves 8--11\% more tasks than greedy. This suggests MCTS is most valuable for tasks requiring multi-step optimization chains where greedy converges prematurely to a local optimum.

\begin{table}[ht]
\centering
\small
\caption{fast\_p: fraction of tasks with best speedup $> p$.}
\label{tab:search-strategy-fastp}
\begin{tabular}{@{}lccc@{}}
\toprule
\textbf{Threshold $p$} & \textbf{Greedy} & \textbf{MCTS} & \textbf{$\Delta$} \\
\midrule
$>1\times$ & 1.000 & 1.000 & -- \\
$>1.5\times$ & 0.649 & \textbf{0.757} & +10.8\% \\
$>2\times$ & 0.595 & \textbf{0.703} & +10.8\% \\
$>3\times$ & 0.568 & \textbf{0.649} & +8.1\% \\
$>5\times$ & 0.486 & \textbf{0.595} & +10.8\% \\
$>10\times$ & 0.189 & \textbf{0.243} & +5.4\% \\
\bottomrule
\end{tabular}
\end{table}

The largest MCTS advantages occur on tasks requiring exploration to escape local optima: L1 task~7 (10.0$\times$ ratio: MCTS 25.0$\times$ vs greedy 2.5$\times$), L2 task~33 (8.0$\times$ ratio: MCTS 8.0$\times$ vs greedy 1.0$\times$), L3 task~31 (6.0$\times$ ratio: MCTS 25.0$\times$ vs greedy 4.2$\times$). In these cases, greedy converges to a correct but suboptimal solution and cannot escape, while MCTS explores alternative subtrees that discover superior optimization paths. Task~46 (Appendix~\ref{appendix:case-l3-46}) demonstrates the extreme case: 43 consecutive failures before the first valid solution---a search depth that tree search's persistence enables while budget-limited single-shot approaches would have long abandoned. Greedy wins on 8 tasks, with the largest margin on L2 task~22 (greedy 9.1$\times$ vs MCTS 1.1$\times$). These tend to be tasks where the optimization landscape is unimodal--depth-first exploitation suffices and MCTS wastes budget on unnecessary exploration.

The results confirm that principled tree search is a significant contributor to \model's performance. The 26\% improvement in geometric mean speedup represents the gap between exploiting a single promising path (greedy) and systematically exploring the optimization landscape (MCTS). Combined with the micro-profiling tool ablation (Appendix~\ref{appendix:kernelbench-ablation}), this establishes that both components--tool-guided feedback and MCTS search--independently and significantly improve optimization quality.

\section{Tool Invocation Mode Ablation}
\label{appendix:tool-mode-ablation}

We conduct a controlled ablation to isolate the contribution of proactive tool orchestration--\model's deterministic execution of all relevant micro-profiling tools before LLM invocation (Section~\ref{sec:profiling-agent}). This ablation tests whether guaranteed comprehensive analysis outperforms the standard reactive pattern where the LLM decides which tools to invoke via function calling.

Both conditions use the full \model micro-profiling tool suite (15 tools), greedy search (30 iterations $\times$ 2 candidates), session memory, and Sonnet 4.6 on A100 GPUs at five temperature settings (0.2, 0.33, 0.45, 0.57, 0.7) with 2 independent rounds per task (10 seeds total). The sole variable is the tool invocation pattern:

\begin{itemize}[nosep]
    \item \textbf{Proactive}: All relevant micro-profiling tools are executed deterministically based on bottleneck classification. Results are injected into the prompt before LLM invocation.
    \item \textbf{Reactive}: The LLM decides which tools to invoke via standard function calling. The same 15 tools are available but invocation is stochastic.
\end{itemize}

Table~\ref{tab:tool-mode-summary} reports aggregate metrics across 42 tasks (best-of-seeds per task). Proactive mode achieves a 23\% higher geometric mean speedup and wins on 10 of 14 decisive tasks ($p = 0.035$). The effect is consistent across difficulty levels (L1, L2, L3).

\begin{table}[ht]
\centering
\small
\caption{Tool invocation mode ablation: Proactive vs Reactive (42 tasks, best-of-seeds).}
\label{tab:tool-mode-summary}
\begin{tabular}{@{}lcc@{}}
\toprule
\textbf{Metric} & \textbf{Reactive} & \textbf{Proactive} \\
\midrule
Geometric mean speedup & 1.12$\times$ & \textbf{1.37$\times$} \\
fast\_p ($>1.5\times$) & 7.1\% & \textbf{19.0\%} \\
fast\_p ($>5\times$) & 2.4\% & \textbf{11.9\%} \\
Task-level wins & 4 & \textbf{10} \\
Wilcoxon $p$ (one-sided) & \multicolumn{2}{c}{0.035*} \\
\bottomrule
\end{tabular}
\end{table}

Table~\ref{tab:tool-mode-fastp} shows the fraction of tasks exceeding progressively higher speedup thresholds. Proactive mode dominates across all thresholds, with the largest gap at moderate speedups where proactive orchestration unlocks optimizations that reactive invocation misses.

\begin{table}[ht]
\centering
\small
\caption{fast\_p by tool invocation mode: fraction of tasks with best speedup $> p$.}
\label{tab:tool-mode-fastp}
\begin{tabular}{@{}lccc@{}}
\toprule
\textbf{Threshold $p$} & \textbf{Reactive} & \textbf{Proactive} & \textbf{$\Delta$} \\
\midrule
$>1\times$ & 0.262 & 0.262 & -- \\
$>1.5\times$ & 0.071 & \textbf{0.190} & +11.9\% \\
$>2\times$ & 0.071 & \textbf{0.167} & +9.5\% \\
$>3\times$ & 0.048 & \textbf{0.119} & +7.1\% \\
$>5\times$ & 0.024 & \textbf{0.119} & +9.5\% \\
$>10\times$ & 0.000 & \textbf{0.071} & +7.1\% \\
\bottomrule
\end{tabular}
\end{table}

The results validate the proactive design choice described in Section~\ref{sec:profiling-agent}. Under reactive invocation, the LLM calls an average of 3--4 tools per iteration, missing critical analyses that would reveal optimization opportunities. Proactive orchestration guarantees that all relevant tools execute, providing the comprehensive profiling context that enables higher-quality optimizations. This is particularly important for CUDA optimization where the relevant bottleneck is often non-obvious--an expert systematically checks all metrics rather than guessing which to examine. The Task~41 trajectory (Appendix~\ref{appendix:case-l2-41}) provides a concrete illustration: the tools emit six ranked, actionable directives for the diagnosed bottleneck, and the agent compounds several of them in one step (epilogue fusion through the lower-ranked BF16 switch) for a $2.8\times$ improvement---reactive invocation calling 3--4 tools would surface only the top of this list, forgoing the rest.

Combined with the micro-profiling tool ablation (Appendix~\ref{appendix:kernelbench-ablation}) and search strategy ablation (Appendix~\ref{appendix:search-strategy-ablation}), this establishes that three components--interpreted tool feedback, proactive orchestration, and MCTS search--each independently and significantly improve optimization quality ($p < 0.05$ for all three).

\label{app:nvcc}
%
%
%


\section{Design Component Case Studies: CUTLASS Optimization Trajectories}
\label{appendix:cutlass-case-studies}

To illustrate how \model's design components work together in practice, we present two complementary KernelBench trajectories that produce CUTLASS-based solutions. Level~2 Task~41 demonstrates \emph{deep refinement}---a single decisive architectural choice followed by iterative profiling-driven polishing. Level~3 Task~46 demonstrates \emph{resilient exploration}---sustained search through 43 failed attempts before achieving a working solution, then rapid profiling-guided convergence. Together they showcase how \model's iterative search, micro-profiling tools, and CUTLASS integration enable optimization trajectories that single-shot generation cannot replicate.

\subsection{Level 2 / Task 41: GEMM + BatchNorm + GELU + ReLU}
\label{appendix:case-l2-41}

\paragraph{Problem.} The reference implements a linear layer ($16384 \times 4096 \to 4096$) followed by batch normalization, GELU activation, and ReLU---four separate PyTorch operations producing four kernel launches with intermediate DRAM round-trips. The compute-bound GEMM dominates runtime, but the memory-bound elementwise tail contributes significant overhead at this batch size.

\paragraph{Trajectory overview.} \model explores 41 iterations total, of which 7 produce valid solutions (34 fail compilation or correctness checks). The successful lineage progresses through iterations 9$\to$12$\to$13$\to$26$\to$28$\to$37$\to$40, achieving speedups of $3.91\times \to 4.11\times \to 11.49\times \to 11.73\times \to 12.00\times \to 12.09\times \to 12.10\times$. Final speedup: \textbf{12.1$\times$}.

\paragraph{The breakthrough (iteration 13).} The most significant optimization event occurs at the third successful iteration, where speedup jumps from $4.11\times$ to $11.49\times$---a $2.8\times$ improvement in a single step. The micro-profiling tools diagnose a memory bottleneck (15.88 sectors/request indicating strided loads, 8.7\,GB read traffic dominating runtime, 244 registers/thread limiting occupancy) and rank a set of remedies, the top two of which the agent applies together. It completes the fusion of the trailing batch-normalization and activation into the GEMM epilogue---a custom \texttt{GeluReluAct} functor plus BatchNorm affine folded into the GEMM weights ($\alpha \cdot W$, $\alpha \cdot \text{bias} + \beta$)---eliminating the last standalone elementwise kernel and its round trip through DRAM; and it switches the GEMM from a TF32 path (\texttt{float} inputs, \texttt{GemmShape<128,128,32>}) to BF16 tensor cores (\texttt{cutlass::bfloat16\_t}, \texttt{GemmShape<128,128,16>}), which halves the dominant read traffic and roughly doubles tensor-core throughput. Both changes attack the same diagnosed bottleneck---off-chip data movement---and together account for the leap; the precision switch supplies most of the magnitude, while the fusion removes the residual inter-kernel traffic and is what makes the BF16 path a single end-to-end CUTLASS kernel rather than a library call.

\paragraph{Profiling-driven polish (iterations 26--40).} After the breakthrough, the profiling tools shift from identifying the primary bottleneck to precision tuning. At iteration~26, ncu identifies the bottleneck as compute-bound with 230 registers/thread; the agent explores a larger tile (\texttt{GemmShape<128,256,32>}) to increase arithmetic intensity. At iteration~28, profiling reveals this larger tile underperforms due to wave quantization effects at the problem's dimensions, and the agent reverts to \texttt{GemmShape<128,128,32>} with different warp shapes. Subsequent iterations explore secondary optimizations---fusing the BF16 cast into the epilogue, adjusting pipeline stages, and tuning warp tile shapes (\texttt{64$\times$128$\times$64} vs \texttt{32$\times$64$\times$64})---each guided by specific ncu metrics. The final four iterations collectively contribute $0.61\times$ additional speedup through these incremental refinements.

\paragraph{CUTLASS integration.} The final solution showcases deep CUTLASS customization: a custom \texttt{GeluReluAct} epilogue functor template (handling both scalar \texttt{float} and vectorized \texttt{cutlass::Array<float, N>}), batch normalization folded into GEMM weights at initialization, and BF16 precision with FP32 accumulation. The agent demonstrates understanding of CUTLASS's type system---defining \texttt{LinearCombinationGeneric<GeluReluAct, float, 4, float, float>} with 4-wide vectorized epilogue access---which requires knowledge of CUTLASS's template conventions that static few-shot examples alone would not generalize.

\begin{tcolorbox}[breakable, colback=backcolour, colframe=ifcoheader, colbacktitle=ifcoheader, coltitle=white, title={\textbf{\model\ -- Level 2 Task 41} --- $12.10\times$ Speedup}, fonttitle=\small, left=2pt, right=2pt, top=2pt, bottom=2pt, boxrule=0.5pt]
\begin{lstlisting}[style=cudastyle]
import os
os.environ['TORCH_CUDA_ARCH_LIST'] = '8.0'
import torch
import torch.nn as nn
from torch.utils.cpp_extension import load_inline

cuda_src = r'''
#include <torch/extension.h>
#include <ATen/cuda/CUDAContext.h>
#include <cuda_runtime.h>
#include <cuda_bf16.h>
#include <cutlass/cutlass.h>
#include <cutlass/array.h>
#include <cutlass/numeric_types.h>
#include <cutlass/gemm/device/gemm_universal.h>
#include <cutlass/epilogue/thread/linear_combination_generic.h>
#include <cutlass/gemm/threadblock/threadblock_swizzle.h>
#include <cmath>

// Fast vectorized FP32 -> BF16 cast: 16B loads, 8B stores, processes 4 elements/thread
__global__ __launch_bounds__(256, 4)
void cast_fp32_to_bf16_vec4(const float4* __restrict__ in, __nv_bfloat162* __restrict__ out, int n4) {
    int idx = blockIdx.x * blockDim.x + threadIdx.x;
    int stride = blockDim.x * gridDim.x;
    for (int i = idx; i < n4; i += stride) {
        float4 f = in[i];
        __nv_bfloat162 a = __floats2bfloat162_rn(f.x, f.y);
        __nv_bfloat162 b = __floats2bfloat162_rn(f.z, f.w);
        ((__nv_bfloat162*)out)[i*2 + 0] = a;
        ((__nv_bfloat162*)out)[i*2 + 1] = b;
    }
}

torch::Tensor cast_to_bf16_fast(torch::Tensor x) {
    TORCH_CHECK(x.is_cuda() && x.dtype() == torch::kFloat32, "need fp32 cuda");
    x = x.contiguous();
    auto out = torch::empty(x.sizes(), x.options().dtype(torch::kBFloat16));
    int64_t n = x.numel();
    if (n % 4 != 0 || ((uintptr_t)x.data_ptr() % 16) != 0) {
        return x.to(torch::kBFloat16);
    }
    int n4 = n / 4;
    int block = 256;
    int grid = std::min(2048, (n4 + block - 1) / block);
    auto stream = at::cuda::getCurrentCUDAStream();
    cast_fp32_to_bf16_vec4<<<grid, block, 0, stream>>>(
        reinterpret_cast<const float4*>(x.data_ptr<float>()),
        reinterpret_cast<__nv_bfloat162*>(out.data_ptr()), n4);
    return out;
}

template <typename T> struct GeluReluAct;
template <> struct GeluReluAct<float> {
    static const bool kIsHeavy = true;
    CUTLASS_HOST_DEVICE float operator()(float x) const {
        if (x <= 0.f) return 0.f;
        float x3 = x * x * x;
        float inner = 0.7978845608f * (x + 0.044715f * x3);
        return 0.5f * x * (1.0f + ::tanhf(inner));
    }
};
template <int N> struct GeluReluAct<cutlass::Array<float, N>> {
    static const bool kIsHeavy = true;
    CUTLASS_HOST_DEVICE cutlass::Array<float, N> operator()(
        cutlass::Array<float, N> const& x) const {
        cutlass::Array<float, N> r; GeluReluAct<float> op;
        CUTLASS_PRAGMA_UNROLL for (int i = 0; i < N; ++i) r[i] = op(x[i]);
        return r;
    }
};

using EpilogueOp = cutlass::epilogue::thread::LinearCombinationGeneric<
    GeluReluAct, float, 4, float, float>;

using GemmDefault = cutlass::gemm::device::GemmUniversal<
    cutlass::bfloat16_t, cutlass::layout::RowMajor,
    cutlass::bfloat16_t, cutlass::layout::ColumnMajor,
    float, cutlass::layout::RowMajor, float,
    cutlass::arch::OpClassTensorOp, cutlass::arch::Sm80,
    cutlass::gemm::GemmShape<128,128,32>,
    cutlass::gemm::GemmShape<64,64,32>,
    cutlass::gemm::GemmShape<16,8,16>,
    EpilogueOp,
    cutlass::gemm::threadblock::GemmIdentityThreadblockSwizzle<8>, 4, 8, 8>;

using GemmSmall = cutlass::gemm::device::GemmUniversal<
    cutlass::bfloat16_t, cutlass::layout::RowMajor,
    cutlass::bfloat16_t, cutlass::layout::ColumnMajor,
    float, cutlass::layout::RowMajor, float,
    cutlass::arch::OpClassTensorOp, cutlass::arch::Sm80,
    cutlass::gemm::GemmShape<64,128,64>,
    cutlass::gemm::GemmShape<32,64,64>,
    cutlass::gemm::GemmShape<16,8,16>,
    EpilogueOp,
    cutlass::gemm::threadblock::GemmIdentityThreadblockSwizzle<8>, 4, 8, 8>;

using GemmBig = cutlass::gemm::device::GemmUniversal<
    cutlass::bfloat16_t, cutlass::layout::RowMajor,
    cutlass::bfloat16_t, cutlass::layout::ColumnMajor,
    float, cutlass::layout::RowMajor, float,
    cutlass::arch::OpClassTensorOp, cutlass::arch::Sm80,
    cutlass::gemm::GemmShape<128,256,32>,
    cutlass::gemm::GemmShape<64,64,32>,
    cutlass::gemm::GemmShape<16,8,16>,
    EpilogueOp,
    cutlass::gemm::threadblock::GemmIdentityThreadblockSwizzle<8>, 3, 8, 8>;

torch::Tensor fused_linear_bn_gelu_relu(torch::Tensor x, torch::Tensor W, torch::Tensor b) {
    TORCH_CHECK(x.is_cuda() && W.is_cuda() && b.is_cuda(), "inputs must be CUDA");
    x = x.contiguous(); W = W.contiguous(); b = b.contiguous();
    int M = x.size(0), K = x.size(1), N = W.size(0);
    auto out = torch::empty({M, N}, b.options());

    auto launch = [&](auto gemm_op_tag) -> cutlass::Status {
        using GemmT = typename decltype(gemm_op_tag)::type;
        GemmT gemm_op;
        typename GemmT::Arguments args(
            cutlass::gemm::GemmUniversalMode::kGemm, {M, N, K}, 1, {1.0f, 1.0f},
            reinterpret_cast<cutlass::bfloat16_t*>(x.data_ptr()),
            reinterpret_cast<cutlass::bfloat16_t*>(W.data_ptr()),
            b.data_ptr<float>(), out.data_ptr<float>(),
            (int64_t)M*K, (int64_t)N*K, 0, (int64_t)M*N, K, K, 0, N);
        return gemm_op(args);
    };

    cutlass::Status status;
    bool small_m = (M <= 64), big = (M >= 512) && (N >= 512) && (K >= 512);
    if (small_m) { struct Tag { using type = GemmSmall; } tag; status = launch(tag); }
    else if (big) { struct Tag { using type = GemmBig; } tag; status = launch(tag); }
    else { struct Tag { using type = GemmDefault; } tag; status = launch(tag); }
    TORCH_CHECK(status == cutlass::Status::kSuccess, "CUTLASS GEMM failed");
    return out;
}
'''

cpp_src = ("torch::Tensor fused_linear_bn_gelu_relu(torch::Tensor x, torch::Tensor W, torch::Tensor b);\n"
           "torch::Tensor cast_to_bf16_fast(torch::Tensor x);\n")

mod = load_inline(
    name='fused_linbn_gelurelu_bf16_v5', cpp_sources=cpp_src, cuda_sources=cuda_src,
    functions=['fused_linear_bn_gelu_relu', 'cast_to_bf16_fast'],
    extra_cuda_cflags=['-I/path/to/cutlass/include', '-I/path/to/cutlass/tools/util/include',
        '--expt-relaxed-constexpr', '-std=c++17', '-gencode=arch=compute_80,code=sm_80',
        '-O3', '-DNDEBUG', '--use_fast_math'], verbose=False)

class ModelNew(nn.Module):
    def __init__(self, in_features, out_features):
        super().__init__()
        self.gemm = nn.Linear(in_features, out_features)
        self.batch_norm = nn.BatchNorm1d(out_features)
        self._fused_W_bf16 = None
        self._fused_b_fp32 = None

    def _build_fused(self, device):
        with torch.no_grad():
            gamma = self.batch_norm.weight.to(device).float()
            beta_p = self.batch_norm.bias.to(device).float()
            mean = self.batch_norm.running_mean.to(device).float()
            var = self.batch_norm.running_var.to(device).float()
            scale = gamma / torch.sqrt(var + self.batch_norm.eps)
            W = self.gemm.weight.to(device).float()
            bias = self.gemm.bias.to(device).float()
            self._fused_W_bf16 = (W * scale.unsqueeze(1)).to(torch.bfloat16).contiguous()
            self._fused_b_fp32 = ((bias - mean) * scale + beta_p).contiguous()

    def forward(self, x):
        if self._fused_W_bf16 is None or self._fused_W_bf16.device != x.device:
            self._build_fused(x.device)
        x_bf16 = mod.cast_to_bf16_fast(x.contiguous()) if x.dtype == torch.float32 else x.to(torch.bfloat16)
        return mod.fused_linear_bn_gelu_relu(x_bf16, self._fused_W_bf16, self._fused_b_fp32)
\end{lstlisting}
\end{tcolorbox}

\subsection{Level 3 / Task 46: NetVLAD with Ghost Clusters}
\label{appendix:case-l3-46}

\paragraph{Problem.} The reference implements NetVLAD with ghost clusters: a multi-step pipeline involving matrix multiplication ($2048 \times 512 \to 37$ clusters), batch normalization, softmax, cluster slicing, weighted sum, batched matrix multiplication (residual computation), L2 normalization, and flatten. The operations span both compute-bound (two matrix multiplications) and memory-bound (softmax, normalization, transpose) regimes, making this a challenging heterogeneous optimization target.

\paragraph{Trajectory overview.} \model explores 46 iterations, with only the final 3 producing valid solutions (43 fail compilation or correctness checks). The first valid solution at iteration~42 achieves only $0.75\times$---a \emph{regression} below baseline. Profiling-guided refinement then produces $2.38\times$ at iteration~44 and $2.79\times$ at iteration~45. Final speedup: \textbf{2.79$\times$}.

\paragraph{Resilience through 43 failures.} This trajectory represents the most extreme exploration ratio in all 250 KernelBench tasks: 46 iterations attempted for 3 successes (15.3$\times$ exploration ratio). The 43 failed iterations include incorrect CUTLASS configurations (wrong layouts for the batched GEMM), numerical errors in fused softmax normalization, and compilation failures from incompatible template instantiations. The search strategy's persistence---continuing to generate candidates informed by prior failures---is critical here: a budget-limited single-shot approach would have abandoned this problem long before iteration~42. When iteration~42 finally compiles and runs correctly, it incorporates layout choices (RowMajor for the first GEMM, ColumnMajor for the batched GEMM) that reflect lessons accumulated across the prior failed attempts.

\paragraph{Profiling-driven recovery from regression (iterations 42--45).} The initial valid solution at $0.75\times$ demonstrates a common failure mode: correct CUTLASS compilation that performs worse than PyTorch's optimized defaults due to poor memory access patterns. The micro-profiling tools quantify the problem precisely: 17.86 sectors/request (ideal: 4, indicating ${\sim}4\times$ wasted bandwidth) and L1 hit rate of only 41.55\%. The recommendations are specific and actionable: ``fix memory access pattern to enable coalescing,'' ``use vectorized loads (float4) where layout permits,'' and ``use shared memory staging to batch and reorder accesses.'' At iteration~44, the agent restructures memory access patterns based on this feedback, achieving $2.38\times$---a $3.2\times$ improvement from profiling guidance alone. At iteration~45, ncu identifies shared memory as the new occupancy limiter and recommends reducing per-block shared memory usage. The agent responds with \texttt{[32][33]} bank-conflict-free padding on shared memory tiles and \texttt{\_\_launch\_bounds\_\_} annotations, reaching the final $2.79\times$.

\paragraph{CUTLASS integration.} Unlike Task~41's single-GEMM deep customization, Task~46 deploys CUTLASS broadly across a heterogeneous pipeline: two distinct GEMM configurations (128$\times$128$\times$16 for the assignment matrix multiplication and 64$\times$64$\times$16 for the batched residual computation), each with layout-aware configuration matched to the operation's access patterns. These CUTLASS GEMMs serve as building blocks alongside hand-written kernels---a fused batch normalization + softmax + slice kernel using warp-shuffle reductions, a tiled 32$\times$32 transpose with bank-conflict padding, and float4 vectorized loads for the normalization passes. This hybrid architecture (CUTLASS for compute-bound GEMMs, hand-tuned CUDA for memory-bound operations) demonstrates \model's ability to match optimization strategy to each operation's bottleneck characteristics.

\begin{tcolorbox}[breakable, colback=backcolour, colframe=ifcoheader, colbacktitle=ifcoheader, coltitle=white, title={\textbf{\model\ -- Level 3 Task 46} --- $2.79\times$ Speedup}, fonttitle=\small, left=2pt, right=2pt, top=2pt, bottom=2pt, boxrule=0.5pt]
\begin{lstlisting}[style=cudastyle]
import os
os.environ['TORCH_CUDA_ARCH_LIST'] = '8.0'
import math
import torch
import torch.nn as nn
from torch.utils.cpp_extension import load_inline

cuda_src = r'''
#include <torch/extension.h>
#include <ATen/cuda/CUDAContext.h>
#include <cuda_runtime.h>
#include <cutlass/gemm/device/gemm.h>
#include <cutlass/gemm/device/gemm_batched.h>

using GemmTF32 = cutlass::gemm::device::Gemm<
    float, cutlass::layout::RowMajor,
    float, cutlass::layout::RowMajor,
    float, cutlass::layout::RowMajor,
    float, cutlass::arch::OpClassTensorOp, cutlass::arch::Sm80,
    cutlass::gemm::GemmShape<128, 128, 16>,
    cutlass::gemm::GemmShape<64, 64, 16>,
    cutlass::gemm::GemmShape<16, 8, 8>>;

using GemmBatchedTF32 = cutlass::gemm::device::GemmBatched<
    float, cutlass::layout::ColumnMajor,
    float, cutlass::layout::RowMajor,
    float, cutlass::layout::RowMajor,
    float, cutlass::arch::OpClassTensorOp, cutlass::arch::Sm80,
    cutlass::gemm::GemmShape<64, 64, 16>,
    cutlass::gemm::GemmShape<32, 32, 16>,
    cutlass::gemm::GemmShape<16, 8, 8>>;

torch::Tensor gemm_rr(torch::Tensor A, torch::Tensor B) {
    int M = A.size(0), K = A.size(1), N = B.size(1);
    auto C = torch::empty({M, N}, A.options());
    GemmTF32 gemm;
    cudaStream_t stream = at::cuda::getCurrentCUDAStream();
    GemmTF32::Arguments args({M,N,K},
        {A.data_ptr<float>(), K}, {B.data_ptr<float>(), N},
        {C.data_ptr<float>(), N}, {C.data_ptr<float>(), N}, {1.0f, 0.0f});
    gemm(args, nullptr, stream);
    return C;
}

torch::Tensor gemm_batched_cr(torch::Tensor A, torch::Tensor B,
                              int Bsz, int M, int N_inner, int N_out) {
    auto C = torch::empty({Bsz, M, N_out}, B.options());
    GemmBatchedTF32 gemm;
    cudaStream_t stream = at::cuda::getCurrentCUDAStream();
    GemmBatchedTF32::Arguments args({M, N_out, N_inner},
        {A.data_ptr<float>(), M}, (int64_t)N_inner * M,
        {B.data_ptr<float>(), N_out}, (int64_t)N_inner * N_out,
        {C.data_ptr<float>(), N_out}, (int64_t)M * N_out,
        {C.data_ptr<float>(), N_out}, (int64_t)M * N_out,
        {1.0f, 0.0f}, Bsz);
    gemm(args, nullptr, stream);
    return C;
}

__global__ void bn_softmax_slice_kernel(
    const float* __restrict__ logits, const float* __restrict__ scale,
    const float* __restrict__ bias, float* __restrict__ out,
    int rows, int K_total, int K) {
    int row = blockIdx.x;
    int tid = threadIdx.x;
    extern __shared__ float sdata[];
    float local_max = -INFINITY;
    for (int i = tid; i < K_total; i += blockDim.x) {
        float v = logits[row * K_total + i] * scale[i] + bias[i];
        sdata[i] = v;
        local_max = fmaxf(local_max, v);
    }
    __syncthreads();
    int lane = tid & 31; int wid = tid >> 5;
    int nwarps = (blockDim.x + 31) >> 5;
    for (int off = 16; off > 0; off >>= 1)
        local_max = fmaxf(local_max, __shfl_xor_sync(0xffffffff, local_max, off));
    __shared__ float warp_max[32]; __shared__ float s_max;
    if (lane == 0) warp_max[wid] = local_max;
    __syncthreads();
    if (wid == 0) {
        float v = (lane < nwarps) ? warp_max[lane] : -INFINITY;
        for (int off = 16; off > 0; off >>= 1) v = fmaxf(v, __shfl_xor_sync(0xffffffff, v, off));
        if (lane == 0) s_max = v;
    }
    __syncthreads();
    float local_sum = 0.f;
    for (int i = tid; i < K_total; i += blockDim.x) {
        float e = __expf(sdata[i] - s_max);
        sdata[i] = e; local_sum += e;
    }
    __syncthreads();
    for (int off = 16; off > 0; off >>= 1)
        local_sum += __shfl_xor_sync(0xffffffff, local_sum, off);
    __shared__ float warp_sum[32]; __shared__ float s_sum;
    if (lane == 0) warp_sum[wid] = local_sum;
    __syncthreads();
    if (wid == 0) {
        float v = (lane < nwarps) ? warp_sum[lane] : 0.f;
        for (int off = 16; off > 0; off >>= 1) v += __shfl_xor_sync(0xffffffff, v, off);
        if (lane == 0) s_sum = v;
    }
    __syncthreads();
    float inv = 1.f / s_sum;
    for (int i = tid; i < K; i += blockDim.x) out[row * K + i] = sdata[i] * inv;
}

torch::Tensor bn_softmax_slice(torch::Tensor logits, torch::Tensor scale, torch::Tensor bias, int K) {
    int rows = logits.size(0), K_total = logits.size(1);
    auto out = torch::empty({rows, K}, logits.options());
    bn_softmax_slice_kernel<<<rows, 128, K_total * sizeof(float),
        at::cuda::getCurrentCUDAStream()>>>(
        logits.data_ptr<float>(), scale.data_ptr<float>(), bias.data_ptr<float>(),
        out.data_ptr<float>(), rows, K_total, K);
    return out;
}

__global__ void sum_assignment_kernel_v2(
    const float* __restrict__ assignment, float* __restrict__ a_sum, int N, int K) {
    int b = blockIdx.x; int tx = threadIdx.x; int ty = threadIdx.y; int by = blockDim.y;
    int base = b * N * K;
    float s = 0.f;
    for (int n = ty; n < N; n += by) s += assignment[base + n * K + tx];
    extern __shared__ float ssum[];
    ssum[ty * K + tx] = s;
    __syncthreads();
    if (ty == 0) {
        float total = 0.f;
        for (int j = 0; j < by; j++) total += ssum[j * K + tx];
        a_sum[b * K + tx] = total;
    }
}

torch::Tensor sum_assignment(torch::Tensor assignment, int Bsz, int N, int K) {
    auto a_sum = torch::empty({Bsz, K}, assignment.options());
    int by = 256 / K; if (by < 1) by = 1; if (by > 32) by = 32;
    dim3 block(K, by);
    sum_assignment_kernel_v2<<<Bsz, block, (size_t)K*by*sizeof(float),
        at::cuda::getCurrentCUDAStream()>>>(
        assignment.data_ptr<float>(), a_sum.data_ptr<float>(), N, K);
    return a_sum;
}

__global__ __launch_bounds__(128, 8) void compute_inv_kernel(
    const float* __restrict__ vlad, const float* __restrict__ a_sum,
    const float* __restrict__ clusters2_t, float* __restrict__ inv_out, int K, int D) {
    int b = blockIdx.x; int tid = threadIdx.x; int bs = blockDim.x;
    extern __shared__ float s_asum[];
    for (int i = tid; i < K; i += bs) s_asum[i] = a_sum[b * K + i];
    __syncthreads();
    int b_off = b * K * D; int D4 = D >> 2;
    const float4* vlad4 = reinterpret_cast<const float4*>(vlad + b_off);
    const float4* c4 = reinterpret_cast<const float4*>(clusters2_t);
    float sumsq = 0.f;
    for (int k = 0; k < K; k++) {
        float as = s_asum[k];
        for (int d = tid; d < D4; d += bs) {
            float4 v4 = __ldg(&vlad4[k*D4 + d]);
            float4 cv = __ldg(&c4[k*D4 + d]);
            float a0=v4.x-as*cv.x, a1=v4.y-as*cv.y, a2=v4.z-as*cv.z, a3=v4.w-as*cv.w;
            sumsq += a0*a0 + a1*a1 + a2*a2 + a3*a3;
        }
    }
    int lane = tid & 31; int wid = tid >> 5; int nwarps = (bs+31)>>5;
    for (int off=16; off>0; off>>=1) sumsq += __shfl_xor_sync(0xffffffff, sumsq, off);
    __shared__ float warp_s[8];
    if (lane == 0) warp_s[wid] = sumsq;
    __syncthreads();
    if (wid == 0) {
        float v = (lane < nwarps) ? warp_s[lane] : 0.f;
        for (int off=16; off>0; off>>=1) v += __shfl_xor_sync(0xffffffff, v, off);
        if (lane == 0) inv_out[b] = rsqrtf(fmaxf(v, 1e-24f));
    }
}

__global__ __launch_bounds__(256, 6) void transpose_normalize_kernel(
    const float* __restrict__ vlad, const float* __restrict__ a_sum,
    const float* __restrict__ clusters2_t, const float* __restrict__ inv_arr,
    float* __restrict__ out, int K, int D) {
    int b = blockIdx.z; int tile_k = blockIdx.x; int tile_d = blockIdx.y;
    int tx = threadIdx.x; int ty = threadIdx.y;
    int k0 = tile_k * 32; int d0 = tile_d * 32;
    __shared__ float tile[32][33];
    __shared__ float s_asum[32];
    if (ty == 0) { int k = k0+tx; s_asum[tx] = (k<K) ? a_sum[b*K+k] : 0.f; }
    __syncthreads();
    float inv = inv_arr[b];
    #pragma unroll
    for (int r = 0; r < 4; r++) {
        int k_local = ty*4+r, d_local = tx, k = k0+k_local, d = d0+d_local;
        float val = 0.f;
        if (k < K && d < D) {
            val = __ldg(&vlad[b*K*D + k*D + d]) - s_asum[k_local]*__ldg(&clusters2_t[k*D+d]);
        }
        tile[k_local][d_local] = val;
    }
    __syncthreads();
    #pragma unroll
    for (int r = 0; r < 4; r++) {
        int d_local = ty*4+r, k_local = tx, d = d0+d_local, k = k0+k_local;
        if (d < D && k < K) out[b*D*K + d*K + k] = tile[k_local][d_local] * inv;
    }
}

torch::Tensor finalize(torch::Tensor vlad, torch::Tensor a_sum,
                       torch::Tensor clusters2_t, int Bsz, int K, int D) {
    auto out = torch::empty({Bsz, D*K}, vlad.options());
    auto inv = torch::empty({Bsz}, vlad.options());
    cudaStream_t stream = at::cuda::getCurrentCUDAStream();
    compute_inv_kernel<<<Bsz, 128, K*sizeof(float), stream>>>(
        vlad.data_ptr<float>(), a_sum.data_ptr<float>(),
        clusters2_t.data_ptr<float>(), inv.data_ptr<float>(), K, D);
    dim3 grid((K+31)/32, (D+31)/32, Bsz); dim3 block(32, 8);
    transpose_normalize_kernel<<<grid, block, 0, stream>>>(
        vlad.data_ptr<float>(), a_sum.data_ptr<float>(),
        clusters2_t.data_ptr<float>(), inv.data_ptr<float>(),
        out.data_ptr<float>(), K, D);
    return out;
}
'''

cpp_src = r'''
torch::Tensor gemm_rr(torch::Tensor A, torch::Tensor B);
torch::Tensor gemm_batched_cr(torch::Tensor A, torch::Tensor B, int Bsz, int M, int N_inner, int N_out);
torch::Tensor bn_softmax_slice(torch::Tensor logits, torch::Tensor scale, torch::Tensor bias, int K);
torch::Tensor sum_assignment(torch::Tensor assignment, int Bsz, int N, int K);
torch::Tensor finalize(torch::Tensor vlad, torch::Tensor a_sum, torch::Tensor clusters2_t, int Bsz, int K, int D);
'''

mod = load_inline(
    name='netvlad_opt3', cpp_sources=cpp_src, cuda_sources=cuda_src,
    functions=['gemm_rr','gemm_batched_cr','bn_softmax_slice','sum_assignment','finalize'],
    extra_cuda_cflags=['-I/path/to/cutlass/include', '-I/path/to/cutlass/tools/util/include',
        '--expt-relaxed-constexpr', '-std=c++17', '-gencode=arch=compute_80,code=sm_80',
        '-O3', '-DNDEBUG', '--use_fast_math'], verbose=False)

class ModelNew(nn.Module):
    def __init__(self, cluster_size, feature_size, ghost_clusters):
        super(ModelNew, self).__init__()
        self.feature_size = feature_size
        self.cluster_size = cluster_size
        init_sc = (1 / math.sqrt(feature_size))
        clusters = cluster_size + ghost_clusters
        self.clusters = nn.Parameter(init_sc * torch.randn(feature_size, clusters))
        self.batch_norm = nn.BatchNorm1d(clusters)
        self.clusters2 = nn.Parameter(init_sc * torch.randn(1, feature_size, cluster_size))
        self.out_dim = self.cluster_size * feature_size
        self._cached = False

    def _build_cache(self, device, dtype):
        bn = self.batch_norm
        scale = (bn.weight / torch.sqrt(bn.running_var + bn.eps)).to(device=device, dtype=dtype).contiguous()
        bias = (bn.bias - bn.running_mean * scale).to(device=device, dtype=dtype).contiguous()
        self.register_buffer('_bn_scale', scale, persistent=False)
        self.register_buffer('_bn_bias', bias, persistent=False)
        self.register_buffer('_clusters_c', self.clusters.detach().contiguous(), persistent=False)
        c2_2d = self.clusters2.detach().view(self.feature_size, self.cluster_size)
        self.register_buffer('_clusters2_t', c2_2d.t().contiguous(), persistent=False)
        self._cached = True

    def forward(self, x, mask=None):
        if not self._cached or self._bn_scale.device != x.device:
            self._build_cache(x.device, x.dtype)
        B, N, D = x.size(0), x.size(1), self.feature_size
        K = self.cluster_size
        x_flat = x.contiguous().view(-1, D)
        logits = mod.gemm_rr(x_flat, self._clusters_c)
        assignment = mod.bn_softmax_slice(logits, self._bn_scale, self._bn_bias, K)
        a_sum = mod.sum_assignment(assignment, B, N, K)
        vlad = mod.gemm_batched_cr(assignment, x.contiguous(), B, K, N, D)
        return mod.finalize(vlad, a_sum, self._clusters2_t, B, K, D)
\end{lstlisting}
\end{tcolorbox}

\section{VeOmni \texttt{dW1} Generated Kernel (Full Source)}
\label{appendix:veomni-dw1-source}

\paragraph{What the kernel computes.}
In MoE training, each token is routed to a few of $E$ experts and each expert applies
its own FFN. VeOmni packs all routed tokens into a single contiguous buffer ordered by
expert and runs a \emph{grouped} GEMM, where the per-expert row counts are encoded as a
cumulative-sum vector \texttt{cumsum\_K}. The \texttt{dW1} kernel is the backward-pass
\emph{weight gradient} of the first FFN projection (\texttt{gate\_up\_proj}): for each
expert $e$ it accumulates $\mathrm{dW1}_e = X_e^{\top} G_e$ over that expert's token
rows, where $X_e$ are the expert's inputs and $G_e$ the gradient of its pre-activation.
The contraction length $K_e$ (the number of tokens routed to expert $e$) therefore
varies per expert and is the only ragged dimension; $M{=}2880$ and $N{=}5760$ are fixed
by the GPT-OSS-120B hidden/intermediate sizes.

\paragraph{Estimating the token-to-expert distribution.}
Because the speedup depends on the per-expert load (the $K_e$), the benchmark must feed
realistic routing imbalance rather than a uniform split. We measured GPT-OSS-120B
routing on $30$ LongBench-v2~\citep{bai2024longbench2} samples $\times$ $36$ MoE layers, recomputing each layer's
top-$K$ assignment from the router logits and counting tokens per expert. The
per-expert counts are heavily skewed within a single (sample, layer)---coefficient of
variation $\mathrm{CV}\approx2.13$, versus $\mathrm{CV}\approx\sqrt{(E-1)/R}\approx0.011$
for a uniform split---and balance out only when pooled across inputs and layers
($\mathrm{CV}\approx0.277$). Fitting generators to the empirical sorted-load curve by
$L_1$ distance, a Dirichlet--Multinomial fits best ($L_1{=}0.129$, vs.\ $0.147$ for a
logistic-normal). We therefore sample each call's routing as
\[
  p \sim \mathrm{Dirichlet}(\alpha\,\mathbf{1}_E),\quad \alpha{=}0.24,\ E{=}128;
  \qquad (K_1,\dots,K_E) \sim \mathrm{Multinomial}(R,\,p),
\]
with $R{=}57{,}768$ total routed rows. The small $\alpha$ reproduces the measured heavy
tail---a few hot experts, a long warm shoulder, and several empty experts
($K_e{=}0$)---exercising the short-tail and empty-expert cases a uniform-shape GEMM never
encounters.

\paragraph{The generated kernel.}
The box below lists the complete \model-generated CUDA source
for the winning \texttt{dW1} kernel (Section~\ref{sec:veomni}), which attains
$1.23\times$ over the expert-tuned Triton baseline on H100. \model's Programmer Agent
generated it at the CuTe source level---composing copy atoms, an MMA atom, and swizzled
layouts into a novel kernel rather than instantiating a high-level CUTLASS GEMM template
or calling a prebuilt library (no cuBLAS, no \texttt{cutlass::gemm::device}). The
only GEMM primitive is \texttt{cute::gemm} over a single
\texttt{SM90\_64x128x16\_F32F16F16\_SS} warp-group MMA atom, fed by a four-stage
\texttt{cp.async} shared-memory pipeline with \texttt{GMMA::Layout\_MN\_SW128\_Atom}
swizzling; the ragged per-expert contraction $K_e$ is handled by CuTe layout-algebra
predication (\texttt{make\_identity\_tensor} + \texttt{copy\_if}), including the
empty-expert ($K_e{=}0$) early-out. The MMA atom is selected through a \texttt{MmaSel}
template specialized for both \texttt{half\_t} and \texttt{bfloat16\_t} (one reusable
kernel, not a one-off), and a \emph{vectorized SMEM-staged epilogue} restages
accumulators into swizzled shared memory (\texttt{GMMA::Layout\_K\_SW128\_Atom}) and
stores them with predicated 128-bit \texttt{UniversalCopy<uint128\_t>} writes to coalesce
the output. This is direct evidence for \model's native raw-CUDA\,+\,CuTe
source-level generation (Contribution~4): the kernel is composed from CuTe primitives,
not produced by template instantiation or library calls. The
listing shows only the device kernel and
its launcher; the (boilerplate) PyTorch binding is omitted.

\begin{tcolorbox}[breakable, colback=backcolour, colframe=ifcoheader, colbacktitle=ifcoheader, coltitle=white, title={\textbf{\model\ -- VeOmni \texttt{dW1} Weight-Gradient Grouped GEMM} --- $1.23\times$ Speedup}, fonttitle=\small, left=2pt, right=2pt, top=2pt, bottom=2pt, boxrule=0.5pt]
\begin{lstlisting}[style=cudastyle]
#include <torch/extension.h>
#include <ATen/cuda/CUDAContext.h>
#include <cuda_runtime.h>

#include <cute/tensor.hpp>
#include <cute/atom/mma_atom.hpp>
#include <cute/atom/copy_atom.hpp>
#include <cute/algorithm/gemm.hpp>
#include <cutlass/numeric_types.h>

using namespace cute;

template <class T> struct MmaSel;
template <> struct MmaSel<cutlass::half_t> {
  using type = SM90_64x128x16_F32F16F16_SS<GMMA::Major::MN, GMMA::Major::MN>;
};
template <> struct MmaSel<cutlass::bfloat16_t> {
  using type = SM90_64x128x16_F32BF16BF16_SS<GMMA::Major::MN, GMMA::Major::MN>;
};

template <int BM, int BN, int BK, int NS, class TA, class TB, class TC>
__global__ __launch_bounds__(128, 3)
void dw1_kernel(int M, int N,
                const TA* __restrict__ a,
                const TB* __restrict__ b,
                TC* __restrict__ c,
                const long long* __restrict__ cumsum_K)
{
  int e = blockIdx.z;
  long long k_end   = cumsum_K[e];
  long long k_start = (e == 0) ? 0LL : cumsum_K[e-1];
  int K = (int)(k_end - k_start);

  const TA* A = a + k_start * (long long)M;
  const TB* B = b + k_start * (long long)N;
  TC*       C = c + (long long)e * (long long)M * (long long)N;

  auto cta_tiler = make_shape(Int<BM>{}, Int<BN>{}, Int<BK>{});

  Tensor mA = make_tensor(make_gmem_ptr(A), make_shape(M, K), make_stride(Int<1>{}, M));
  Tensor mB = make_tensor(make_gmem_ptr(B), make_shape(N, K), make_stride(Int<1>{}, N));
  Tensor mC = make_tensor(make_gmem_ptr(C), make_shape(M, N), make_stride(N, Int<1>{}));

  auto cta_coord = make_coord(blockIdx.x, blockIdx.y, _);
  Tensor gA = local_tile(mA, cta_tiler, cta_coord, Step<_1, X, _1>{});
  Tensor gB = local_tile(mB, cta_tiler, cta_coord, Step< X,_1, _1>{});
  Tensor gC = local_tile(mC, cta_tiler, cta_coord, Step<_1,_1,  X>{});

  Tensor cAid = local_tile(make_identity_tensor(make_shape(M, K)), cta_tiler, cta_coord, Step<_1, X, _1>{});
  Tensor cBid = local_tile(make_identity_tensor(make_shape(N, K)), cta_tiler, cta_coord, Step< X,_1, _1>{});
  Tensor cCid = local_tile(make_identity_tensor(make_shape(M, N)), cta_tiler, cta_coord, Step<_1,_1,  X>{});

  auto sA_layout = tile_to_shape(GMMA::Layout_MN_SW128_Atom<TA>{}, make_shape(Int<BM>{}, Int<BK>{}, Int<NS>{}));
  auto sB_layout = tile_to_shape(GMMA::Layout_MN_SW128_Atom<TB>{}, make_shape(Int<BN>{}, Int<BK>{}, Int<NS>{}));

  extern __shared__ char smem_raw[];
  constexpr size_t aBytes = (size_t)BM * BK * NS * sizeof(TA);
  constexpr size_t offB   = ((aBytes + 127) / 128) * 128;
  TA* sAptr = reinterpret_cast<TA*>(smem_raw);
  TB* sBptr = reinterpret_cast<TB*>(smem_raw + offB);

  Tensor sA = make_tensor(make_smem_ptr(sAptr), sA_layout);
  Tensor sB = make_tensor(make_smem_ptr(sBptr), sB_layout);

  using CpA = Copy_Atom<SM80_CP_ASYNC_CACHEGLOBAL_ZFILL<uint128_t>, TA>;
  using CpB = Copy_Atom<SM80_CP_ASYNC_CACHEGLOBAL_ZFILL<uint128_t>, TB>;
  TiledCopy copyA = make_tiled_copy(CpA{}, Layout<Shape<_16,_8>>{}, Layout<Shape<_8,_1>>{});
  TiledCopy copyB = make_tiled_copy(CpB{}, Layout<Shape<_16,_8>>{}, Layout<Shape<_8,_1>>{});

  ThrCopy tca = copyA.get_slice(threadIdx.x);
  ThrCopy tcb = copyB.get_slice(threadIdx.x);
  Tensor tAgA = tca.partition_S(gA);
  Tensor tBgB = tcb.partition_S(gB);
  Tensor sA_w = as_position_independent_swizzle_tensor(sA);
  Tensor sB_w = as_position_independent_swizzle_tensor(sB);
  Tensor tAsA = tca.partition_D(sA_w);
  Tensor tBsB = tcb.partition_D(sB_w);
  Tensor tAcA = tca.partition_S(cAid);
  Tensor tBcB = tcb.partition_S(cBid);

  Tensor tApA = make_tensor<bool>(shape(tAsA(_,_,_,0)));
  Tensor tBpB = make_tensor<bool>(shape(tBsB(_,_,_,0)));

  using Atom = typename MmaSel<TA>::type;
  TiledMMA mma = make_tiled_mma(Atom{});
  ThrMMA thr_mma = mma.get_slice(threadIdx.x);
  Tensor tCsA = thr_mma.partition_A(sA);
  Tensor tCsB = thr_mma.partition_B(sB);
  Tensor tCgC = thr_mma.partition_C(gC);
  Tensor tCrA = thr_mma.make_fragment_A(tCsA);
  Tensor tCrB = thr_mma.make_fragment_B(tCsB);
  Tensor tCrC = thr_mma.make_fragment_C(tCgC);
  clear(tCrC);

  int k_tiles = (K + BK - 1) / BK;

  auto fillA = [&](int kt) {
    Tensor cc = tAcA(_,_,_,kt);
    CUTE_UNROLL
    for (int i = 0; i < size(tApA); ++i) tApA(i) = elem_less(cc(i), make_coord(M, K));
  };
  auto fillB = [&](int kt) {
    Tensor cc = tBcB(_,_,_,kt);
    CUTE_UNROLL
    for (int i = 0; i < size(tBpB); ++i) tBpB(i) = elem_less(cc(i), make_coord(N, K));
  };

  int smem_pipe_write = 0;
  int smem_pipe_read  = 0;

  CUTE_UNROLL
  for (int s = 0; s < NS - 1; ++s) {
    if (s < k_tiles) {
      fillA(s); copy_if(copyA, tApA, tAgA(_,_,_,s), tAsA(_,_,_,smem_pipe_write));
      fillB(s); copy_if(copyB, tBpB, tBgB(_,_,_,s), tBsB(_,_,_,smem_pipe_write));
    }
    cp_async_fence();
    smem_pipe_write = (smem_pipe_write + 1) % NS;
  }

  CUTE_NO_UNROLL
  for (int kt = 0; kt < k_tiles; ++kt) {
    cp_async_wait<NS - 2>();
    __syncthreads();

    int kload = kt + (NS - 1);
    if (kload < k_tiles) {
      fillA(kload); copy_if(copyA, tApA, tAgA(_,_,_,kload), tAsA(_,_,_,smem_pipe_write));
      fillB(kload); copy_if(copyB, tBpB, tBgB(_,_,_,kload), tBsB(_,_,_,smem_pipe_write));
    }
    cp_async_fence();
    smem_pipe_write = (smem_pipe_write + 1) % NS;

    warpgroup_fence_operand(tCrC);
    warpgroup_arrive();
    cute::gemm(mma, tCrA(_,_,_,smem_pipe_read), tCrB(_,_,_,smem_pipe_read), tCrC);
    warpgroup_commit_batch();
    warpgroup_wait<0>();
    warpgroup_fence_operand(tCrC);

    smem_pipe_read = (smem_pipe_read + 1) % NS;
  }

  // ---------- Vectorized SMEM-staged epilogue (coalesced 128-bit stores) ----------
  auto sC_layout = tile_to_shape(GMMA::Layout_K_SW128_Atom<TC>{}, make_shape(Int<BM>{}, Int<BN>{}));
  Tensor sC = make_tensor(make_smem_ptr(reinterpret_cast<TC*>(smem_raw)), sC_layout);

  Tensor tCrC_out = make_tensor<TC>(shape(tCrC));
  CUTE_UNROLL
  for (int i = 0; i < size(tCrC); ++i) tCrC_out(i) = static_cast<TC>(float(tCrC(i)));

  Tensor tCsC = thr_mma.partition_C(sC);
  __syncthreads();
  copy(tCrC_out, tCsC);
  __syncthreads();

  using CpC = Copy_Atom<UniversalCopy<uint128_t>, TC>;
  TiledCopy copyC = make_tiled_copy(CpC{},
      Layout<Shape<_8,_16>, Stride<_16,_1>>{},
      Layout<Shape<_1,_8>>{});
  ThrCopy tcc = copyC.get_slice(threadIdx.x);
  Tensor tCsC2 = tcc.partition_S(sC);
  Tensor tCgC2 = tcc.partition_D(gC);
  Tensor tCcC2 = tcc.partition_D(cCid);
  Tensor tCpC  = make_tensor<bool>(shape(tCsC2(_,_,_)));
  CUTE_UNROLL
  for (int i = 0; i < size(tCpC); ++i) tCpC(i) = elem_less(tCcC2(i), make_coord(M, N));
  copy_if(copyC, tCpC, tCsC2, tCgC2);
}

template <class TA, class TB, class TC>
void launch(torch::Tensor a, torch::Tensor b, torch::Tensor c, torch::Tensor cumsum_K,
            int M, int N, int E) {
  constexpr int BM = 128, BN = 128, BK = 32, NS = 4;
  size_t aBytes = (size_t)BM * BK * NS * sizeof(TA);
  size_t offB   = ((aBytes + 127) / 128) * 128;
  size_t bBytes = (size_t)BN * BK * NS * sizeof(TB);
  size_t mainloop_smem = offB + bBytes;
  size_t epi_smem = (size_t)BM * BN * sizeof(TC);
  int smem = (int)(mainloop_smem > epi_smem ? mainloop_smem : epi_smem);

  dim3 grid((M + BM - 1) / BM, (N + BN - 1) / BN, E);
  dim3 block(128);

  auto kptr = &dw1_kernel<BM, BN, BK, NS, TA, TB, TC>;
  cudaFuncSetAttribute(kptr, cudaFuncAttributeMaxDynamicSharedMemorySize, smem);

  auto stream = at::cuda::getCurrentCUDAStream();
  kptr<<<grid, block, smem, stream>>>(
      M, N,
      reinterpret_cast<const TA*>(a.data_ptr()),
      reinterpret_cast<const TB*>(b.data_ptr()),
      reinterpret_cast<TC*>(c.data_ptr()),
      reinterpret_cast<const long long*>(cumsum_K.data_ptr<int64_t>()));
}

torch::Tensor dw1_forward(torch::Tensor a, torch::Tensor b, torch::Tensor cumsum_K,
                          int64_t num_experts, int64_t M, int64_t N) {
  TORCH_CHECK(a.is_cuda() && b.is_cuda(), "inputs must be CUDA");
  a = a.contiguous();
  b = b.contiguous();
  cumsum_K = cumsum_K.to(torch::kInt64).contiguous();
  int E = (int)num_experts;
  auto out = torch::empty({E, M, N}, a.options());
  if (a.scalar_type() == torch::kHalf) {
    launch<cutlass::half_t, cutlass::half_t, cutlass::half_t>(a, b, out, cumsum_K, (int)M, (int)N, E);
  } else if (a.scalar_type() == torch::kBFloat16) {
    launch<cutlass::bfloat16_t, cutlass::bfloat16_t, cutlass::bfloat16_t>(a, b, out, cumsum_K, (int)M, (int)N, E);
  } else {
    TORCH_CHECK(false, "unsupported dtype");
  }
  return out;
}
\end{lstlisting}
\end{tcolorbox}

\section{Energy-Aware Optimization: Methodology}
\label{appendix:energy-methodology}

This appendix expands the energy-aware extension of
Section~\ref{sec:energy-extension}: how \model measures per-kernel energy
(\S\ref{app:nvml}), how energy enters the search as a secondary objective
(\S\ref{app:secondary}), the energy-only micro-profiling tools added
(\S\ref{app:energy-tools}), and the literature on dominant energy-heavy GPU
bottlenecks with the \model tool targeting each (\S\ref{app:energy-lit}).

\subsection{NVML Millijoule Measurement Protocol}
\label{app:nvml}

\model measures real GPU energy via NVIDIA's on-board counter
\texttt{nvmlDeviceGet\allowbreak Total\allowbreak Energy\allowbreak Consumption()}, a monotonic millijoule accumulator and
the only direct per-device energy counter NVIDIA exposes. Naive use is
unreliable: on A100/H100 the on-board power sensor samples only $\sim$25\% of the runtime,
leaving the rest unmonitored, so accounting for this duty-cycling can change measured
energy by an average of $35\%$ (up to $65\%$) relative to an external power
meter~\citep{yang2024parttime}. We therefore measure under a controlled protocol
whenever we report millijoules. We lock GPU clocks (e.g.\ 1410\,MHz on A100) to remove
DVFS as a confound, warm up to thermal steady state, and use windows long enough
($\sim$2\,s) to take a stable median over 30 repetitions. To recover the kernel's own
consumption we subtract idle draw, reporting \emph{dynamic} energy
$E_\text{dyn} = E_\text{raw} - P_\text{idle}\,t$ ($P_\text{idle}\approx76.7$\,W on our
A100, $t$ the measured runtime), removing the device-wide constant draw NVML always
includes. Crucially, because NVML's per-process baseline drifts, we measure both arms'
kernels back-to-back in a single session.

NVML yields robust \emph{device-total} dynamic energy but cannot attribute it to a single
SM or to one kernel within a fused graph---so during search we rank candidates with the
deterministic counter-based proxy (\S\ref{app:secondary}), reserving this protocol for
post-hoc validation of winning kernels.

\subsection{Energy as a Lexicographic Secondary Objective}
\label{app:secondary}

The reward extension (Eq.~\ref{eq:energy-reward}) places energy under strict
lexicographic priority: $\varepsilon$ is small enough that any speedup gain dominates any
energy gain, so \model never trades speed for energy---energy only differentiates
speed-equivalent candidates. The extension is a single environment switch
(\texttt{IFCO\_ENERGY\_AWARE}) that activates the reward term, an energy-preferences
prompt section, and the energy-only tools; a passive mode (\texttt{IFCO\_ENERGY\_MEASURE})
records energy \emph{without} acting on it, giving a control arm whose search remains
byte-identical to standard \model. Search strategy, tool filtering, profiling pipeline,
and candidate generation are otherwise unchanged.

Because NVML is device-wide and noisy, the \texttt{energy\_reduction} term in the reward
is the \emph{deterministic} picojoule proxy $E_\text{proxy}$ of Eq.~\ref{eq:energy-proxy}
(Section~\ref{sec:energy-extension}), computed from \texttt{ncu} counters \model already
collects and weighted by the per-operation energy hierarchy of
Horowitz~\citep{horowitz2014computing} and AccelWattch~\citep{kandiah2021accelwattch}. The
reduction ratio $E_\text{proxy}(\text{baseline})/E_\text{proxy}(\text{candidate})$ is
intrinsic to the kernel and independent of wall-clock time, so it rewards reductions in
hardware activity that timing alone cannot see.

\subsection{Energy-Only Micro-Profiling Tools}
\label{app:energy-tools}

\model's profiling runs a registry of micro-profiling tools filtered by the Stage-1 speed
bottleneck. The energy extension adds a separate registry partition that bypasses this
filter and always fires (energy waste is relevant regardless of the speed bottleneck),
routing findings to a subordinate ``Energy Efficiency Notes'' prompt section. Each tool
fires only when its pattern is present \emph{and} latency-hidden (the corresponding stall
metric is low)---i.e.\ it targets energy waste that timing does \emph{not} already
penalize, since any pattern that hurt speed is already handled by the speed tools.
Severity is capped below that of speed-critical findings, enforcing the lexicographic
priority. Table~\ref{tab:energy-tools} lists the four tools.

\begin{table*}[t]
\centering
\footnotesize
\caption{Energy-only micro-profiling tools. Each fires only when latency-hidden, so it
targets energy waste invisible to timing.}
\label{tab:energy-tools}
\begin{tabular}{@{}llll@{}}
\toprule
\textbf{Tool} & \textbf{ncu signal} & \textbf{Fires when} & \textbf{Energy mechanism} \\
\midrule
BankConflict     & shared-memory bank conflicts          & low short-scoreboard stall   & $N$-way conflict $=N\times$ SRAM replay \\
SpillTraffic     & local-memory ld/st sectors            & low long-scoreboard stall    & spills ride L1$\to$L2$\to$DRAM \\
Uncoalesced      & sectors-per-request $>1.2\times$ ideal & not memory-bound            & extra DRAM sectors / row activations \\
RedundantBarrier & barriers present                       & $\sim$0 barrier stall       & zero-stall barriers idle the SM \\
\bottomrule
\end{tabular}
\end{table*}

The \texttt{Swish} win of Section~\ref{sec:energy-extension} came from the reward's
instruction-count term (B4) rather than any of these four tools; surfacing that signal
directly at the prompt level---e.g.\ a tool that reports SASS instruction count and flags
IEEE operations replaceable by fast-math intrinsics under a bandwidth-bound roofline---is
a natural extension we leave to future work.

\subsection{Energy-Heavy GPU Bottlenecks and Targeting Tools}
\label{app:energy-lit}

Table~\ref{tab:energy-lit} summarizes the GPU energy-efficiency literature, ranking the
dominant sources of dynamic kernel energy and mapping each to the \model mechanism that
targets it. The dominant \emph{kernel-level} lever is reducing off-chip data movement
(B1): a DRAM access costs $\sim$100--200$\times$ a register access, so data movement, not
arithmetic, dominates energy~\citep{horowitz2014computing,leng2013gpuwattch,hong2010integrated}.
This lever is largely \emph{speed-coupled} (less data movement is also faster), so the speed-only
search already captures most of it; the energy-aware objective targets the
\emph{residual}---energy reducible at fixed speed (bank conflicts, redundant barriers,
coalescing/precision that leave wall-clock unchanged). Frequency/DVFS and occupancy, the
largest \emph{speed-independent} levers reported in the literature, are runtime knobs
outside \model's kernel-generation scope.

\begin{table*}[t]
\centering
\scriptsize
\setlength{\tabcolsep}{5pt}
\caption{Dominant sources of dynamic energy in GPU kernels (literature) and the \model
mechanism targeting each. B1--B4 are the proxy terms of \S\ref{app:secondary}.}
\label{tab:energy-lit}
\resizebox{\textwidth}{!}{%
\begin{tabular}{@{}>{\raggedright\arraybackslash}p{3.6cm}>{\raggedright\arraybackslash}p{4.8cm}>{\raggedright\arraybackslash}p{3.5cm}>{\raggedright\arraybackslash}p{4.7cm}@{}}
\toprule
\textbf{Bottleneck} & \textbf{Impact} & \textbf{Reference} & \textbf{\model mechanism} \\
\midrule
Off-chip DRAM data movement     & DRAM $\sim$100--200$\times$ register/compute    & \citealp{horowitz2014computing,leng2013gpuwattch,hong2010integrated} & B1; coalescing/\allowbreak vectorization tools \\
Register file / spills          & register file largest dynamic-power component   & \citet{kandiah2021accelwattch}     & B3; SpillTraffic tool \\
Uncoalesced global access       & extra 32\,B sectors $\to$ extra DRAM row activations & \citet{chatterjee2017architecting} & B2; Uncoalesced tool \\
Redundant compute / precision   & per-FLOP $\sim$1--10\,pJ; decoupled from speed   & \citet{horowitz2014computing}      & B4 \\
Shared-memory bank conflicts    & $N$-way conflict $=N\times$ SRAM replay          & \citet{horowitz2014computing}      & BankConflict tool \\
Over-synchronization / barriers & idle-SM energy at zero-stall barriers            & \citet{nayak2024oversync}          & RedundantBarrier tool \\
\midrule
DVFS / occupancy (out of scope) & largest speed-independent lever (30--79\%)       & \citet{hong2010integrated}         & runtime knobs; not code-gen \\
\bottomrule
\end{tabular}}
\end{table*}

\end{document}